\documentclass[10pt,twocolumn,letterpaper]{article}

\usepackage[pagenumbers]{cvpr} % To force page numbers, e.g. for an arXiv version
\usepackage{appendix}  % 附录包
\usepackage{listings}  % 代码展示
\usepackage{xcolor}    % 颜色支持
\usepackage[skins]{tcolorbox} % 文本框
\usepackage{booktabs}
\usepackage{multirow} % 用于 \multirow 命令
\usepackage{multicol}
\usepackage{enumitem} % 用于调整列表间距
\usepackage{colortbl}
\usepackage{adjustbox}  % 导言区
\usepackage{pifont}   % 提供 \ding{51} (✓) 和 \ding{55} (✗)
\usepackage[T1]{fontenc}
\usepackage{subcaption}
\usepackage{graphicx}

\newtcolorbox{SplitBox}[1][]{%
    enhanced,                % 开启高级模式 (画虚线必须)
    arc=3pt,                 % 圆角大小
    boxrule=1.5pt,           % 边框粗细
    segmentation style={
        dashed,              % 样式：虚线
        gray!60,             % 颜色：灰色
        line width=0.8pt     % 粗细
    },
    middle=5pt,              % 分割线附近的垂直间距
    top=5pt, bottom=5pt, left=5pt, right=5pt,
    drop shadow,
    rounded corners,
    #1
}

\definecolor{codevarcolor}{HTML}{6F42C1} % VS Code 风格紫
\newcommand{\codevar}[1]{\textbf{\textcolor{codevarcolor}{#1}}}

\usepackage{listings}
\usepackage{xcolor}

\definecolor{jsonkey}{HTML}{0000CD} % 键名 (深蓝)
\definecolor{jsonval}{HTML}{008000} % 字符串值 (绿色)
\definecolor{jsonnum}{HTML}{FF0000} % 数字 (红色)
\definecolor{backcolour}{HTML}{FDFBF6} % 背景色 (和你之前的 Box 背景一致)

\lstdefinelanguage{json}{
    basicstyle=\scriptsize\ttfamily, % 字体大小和样式
    numbers=left,                    % 行号在左边 (不需要可改为 none)
    numberstyle=\tiny\color{gray},   % 行号样式
    stepnumber=1,
    numbersep=8pt,
    showstringspaces=false,          % 不显示空格标记
    breaklines=true,                 % 自动换行
    frame=none,                      % 取消自带边框 (因为外面有 SplitBox 了)
    backgroundcolor=\color{backcolour}, % 背景色
    literate=
     *{0}{{{\color{jsonnum}0}}}{1}
      {1}{{{\color{jsonnum}1}}}{1}
      {2}{{{\color{jsonnum}2}}}{1}
      {3}{{{\color{jsonnum}3}}}{1}
      {4}{{{\color{jsonnum}4}}}{1}
      {5}{{{\color{jsonnum}5}}}{1}
      {6}{{{\color{jsonnum}6}}}{1}
      {7}{{{\color{jsonnum}7}}}{1}
      {8}{{{\color{jsonnum}8}}}{1}
      {9}{{{\color{jsonnum}9}}}{1}
      {:}{{{\color{black}{:}}}}{1}
      {,}{{{\color{black}{,}}}}{1}
      {\{}{{{\color{black}{\{}}}}{1}
      {\}}{{{\color{black}{\}}}}}{1}
      {[}{{{\color{black}{[}}}}{1}
      {]}{{{\color{black}{]}}}}{1},
    stringstyle=\color{jsonval},     % 字符串默认绿色
    commentstyle=\color{gray},
    morestring=[b]",
    morecomment=[l]{//},
}

\makeatletter
\lst@AddToHook{OutputOther}{\@dded}
\def\@dded{
  \ifnum\lst@mode=\lst@Pmode
    \def\lst@thestyle{\color{jsonkey}}
  \fi
}
\makeatother

\definecolor{cvprblue}{rgb}{0.21,0.49,0.74}
\definecolor{gptpurple}{HTML}{6F42C1} % 紫色边框
\definecolor{gptpurpleback}{HTML}{F6F2FF} % 极淡的淡紫色背景

\usepackage[pagebackref,breaklinks,colorlinks,allcolors=cvprblue]{hyperref}
\usepackage[capitalize]{cleveref}

\def\paperID{29467} % *** Enter the Paper ID here
\def\confName{CVPR}
\def\confYear{2026}

\title{Paper2SysArch: Structure‑Constrained System Architecture Generation from Scientific Papers}

\author{Ziyi Guo\\
Beijing University of Posts and Telecommunications\\
Beijing, China\\
{\tt\small guoziyi2023@bupt.edu.cn}
\and
Zhou Liu\\
Peking University\\
Beijing, China\\
{\tt\small zhouliu25@stu.pku.edu.cn}
\and
Wentao Zhang\\
Peking University\\
Beijing, China\\
{\tt\small wentao.zhang@pku.edu.cn}
}

\begin{document}
\maketitle

%%%%%%%%% ABSTRACT
\begin{abstract}
% \textcolor{red}{Mock!}
% System architecture diagrams are essential for understanding scientific papers, yet their creation remains largely manual and subjective. We introduce \textbf{Paper2SysArch}, the first large-scale benchmark for automated architecture diagram generation from papers, comprising 3,000 paper-diagram pairs with structured ground truth. Unlike pixel-based image generation, our benchmark focuses on semantic structure, providing complete hierarchical graph annotations. We propose a three-tier evaluation framework assessing semantic accuracy, layout quality, and visual clarity. As a strong baseline, we develop an agent-based system achieving 65.3\% overall score. Our benchmark enables reproducible research in this important but understudied domain.
The manual creation of system architecture diagrams for scientific papers is a time-consuming and subjective process, while existing generative models lack the necessary structural control and semantic understanding for this task. A primary obstacle hindering research and development in this domain has been the profound lack of a standardized benchmark to quantitatively evaluate the automated generation of diagrams from text. To address this critical gap, we introduce a novel and comprehensive benchmark, the first of its kind, designed to catalyze progress in automated scientific visualization. It consists of \textit{3,000} research papers paired with their corresponding high-quality ground-truth diagrams and is accompanied by a three-tiered evaluation metric assessing semantic accuracy, layout coherence, and visual quality. Furthermore, to establish a strong baseline on this new benchmark, we propose Paper2SysArch, an end-to-end system that leverages multi-agent collaboration to convert papers into structured, editable diagrams. To validate its performance on complex cases, the system was evaluated on a manually curated and more challenging subset of these papers, where it achieves a composite score of \textit{69.0}. This work's principal contribution is the establishment of a large-scale, foundational benchmark to enable reproducible research and fair comparison. Meanwhile, our proposed system serves as a viable proof-of-concept, demonstrating a promising path forward for this complex task. 
\end{abstract}

%%%%%%%%% BODY TEXT
\section{Introduction}
\label{sec:intro}

\begin{figure}[ht!]
    \centering
    \includegraphics[width=1.0\linewidth]{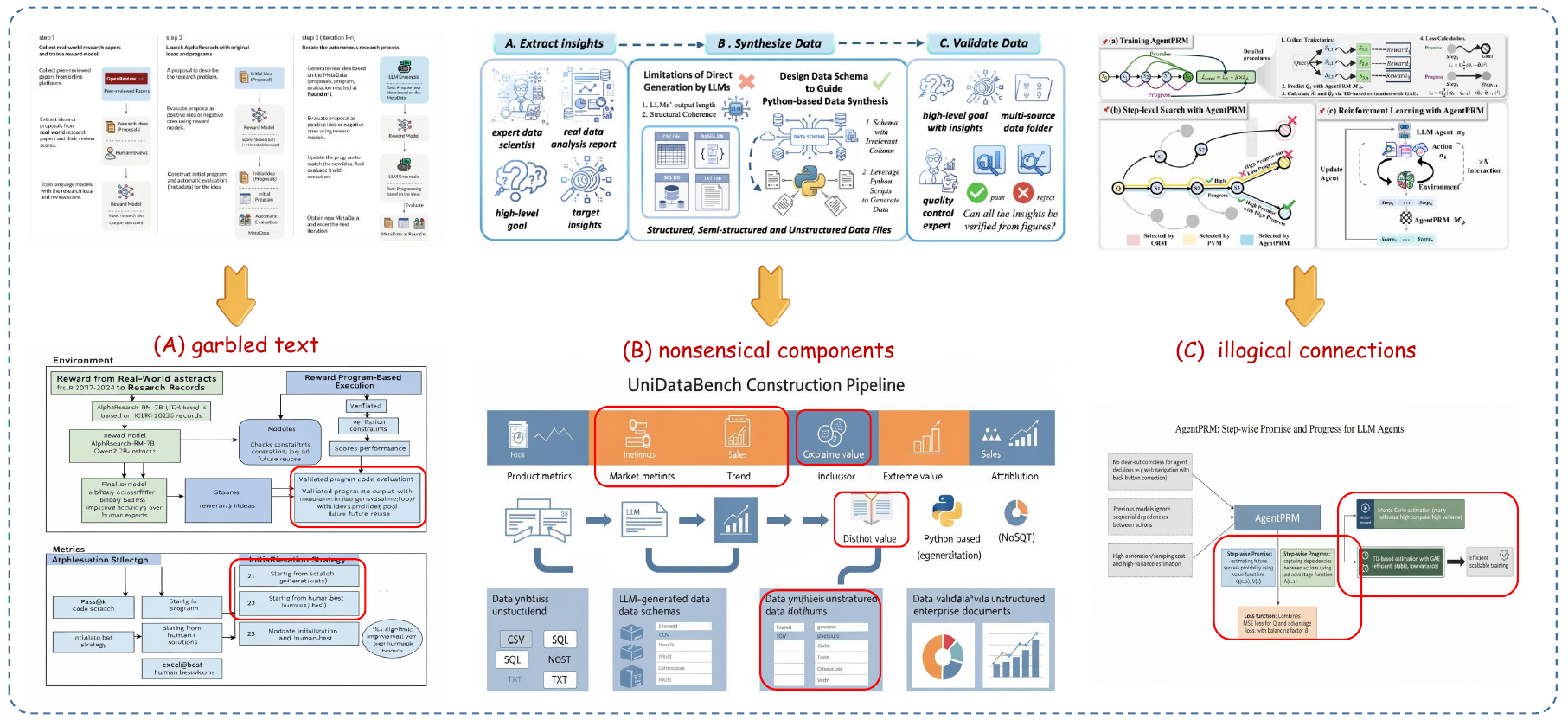}
    \caption{Typical failure cases of general-purpose image generation models for scientific diagrams (evaluated using the Nanobanana \cite{gemini} model).}
    \label{fig:bad_cases}
    \vspace{-10pt}
\end{figure}

System architecture diagrams are pivotal for the dissemination of scientific research. They serve as condensed visual representations, intuitively illustrating a method's key components, data flow, and core logic. A clear and accurate diagram not only helps readers quickly grasp the essence of a paper but also reflects the researcher's deep understanding of system hierarchy and abstraction. However, the creation of such diagrams is currently a largely manual process \cite{chen2024viseval, chen2024textdiffuser}. Researchers must painstakingly extract core concepts from lengthy texts, organize them into logical modules, and meticulously design the layout and style—a process that is time-consuming, subjective, and lacks reproducibility \cite{chen2024opportunities}.

Concurrently, modern image generation techniques, such as Diffusion models \cite{nichol2021improved,song2020denoising}, can produce visually compelling graphics but struggle with scientific diagrams that require precise logic, structure, and textual representation \cite{liao2025challenges}. As illustrated in Figure ~\ref{fig:bad_cases}, these general-purpose models tend to generate ``flat" pixel-based images with factual errors (e.g., garbled text and nonsensical components) and chaotic topological structures (e.g., illogical connections), lacking explicit control over diagrammatic elements. Furthermore, their outputs are neither editable nor reusable \cite{hong2023chatdev}. While some tools attempt to generate diagrams from simple instructions, they fail to process long and complex documents like academic papers.

A more fundamental problem is that research progress on the automated generation of \textbf{knowledge-dense, structured visual content} like system architecture diagrams has been slow. A key obstacle is the lack of a standardized and quantifiable benchmark. Without a recognized ``gold standard" and evaluation framework, different methods cannot be compared under fair conditions, making it difficult for the field to measure substantive technological advancements.

To address this critical gap, we make two core contributions. First, we introduce the Paper2SysArch Benchmark, the first large-scale, comprehensive benchmark in this domain. It comprises 3,000 research papers paired with their corresponding high-quality, structured ground-truth diagrams. We also designed an accompanying three-tier automated evaluation framework to quantitatively assess the quality of generated diagrams from three orthogonal perspectives: semantic consistency, layout rationality, and visual clarity.

Second, to establish a baseline on this new benchmark, we propose Paper2SysArch, an end-to-end system based on multi-agent collaboration that automatically converts a full scientific paper into a structured and editable system architecture diagram. Unlike pixel-level generative models, the output of Paper2SysArch is semantically explicit, hierarchically clear, and fully controllable \cite{hong2023chatdev}, charting a new path for automated scientific visualization.

The primary contribution of this work is the establishment of a foundational benchmark to lay the groundwork for future reproducible research and fair comparisons in the field. Our proposed Paper2SysArch system also serves as an effective proof-of-concept, demonstrating that the automated generation of system architecture diagrams from papers is a viable and promising research direction. Our main contributions are summarized as follows:
\begin{itemize}
    \item We introduce the first large-scale ``Paper-to-System-Architecture" benchmark, featuring 3,000 paper-diagram pairs that span multiple domains, scales, and structural types.
    \item We develop a three-tier automated evaluation framework for quantitative assessment from the orthogonal dimensions of semantic consistency, layout rationality, and visual readability, enabling reproducible comparisons.
    \item We design Paper2SysArch, an end-to-end system that leverages multi-agent collaboration to automate the full pipeline of semantic extraction, structural modeling, layout planning, and graphical rendering, producing an editable format that supports subsequent manual refinement.
\end{itemize}

%-------------------------------------------------------------------------
\section{Related Work}
\label{sec:related}
\subsection{Benchmarks for Diagram Generation}
DiagramGenBenchmark \cite{diagramgenbenchmark2024} is a large-scale benchmark aggregated from multiple diagram types, which primarily evaluates a model's ability to generate underlying diagram code (e.g., LaTeX). MagicGeoBench \cite{magicgeobench2024}, in contrast, focuses on generating plane geometry figures for middle school math problems, employing a formal language solver to ensure geometric correctness and relying on user rankings for evaluation. Other efforts have focused more on the evaluation methodology itself. For instance, AI2D-Caption \cite{ai2dcaption2024} extends the AI2D dataset with dense annotations to provide a solid data foundation for text-to-diagram generation tasks. DiagramEval \cite{diagrammergpt2024} proposes an automated evaluation pipeline for diagrams generated from scientific papers. In contrast to these prior works, our proposed Paper2SysArch benchmark possesses several distinct characteristics. It specifically targets the complex and knowledge-intensive task of generating system architecture diagrams from full scientific papers. Compared to DiagramEval, our work offers key advantages: first, we aim to generate editable, structured diagrams, emphasizing downstream practical utility; second, we have constructed a larger benchmark and provide a more comprehensive quality assessment framework that moves beyond singular structural alignment.

\subsection{Generation of Visual Content and Diagrams}
The advent of foundational text-to-image models, such as Stable Diffusion \citep{rombach2022high, podell2023sdxl}, has enabled high-fidelity synthesis of naturalistic images. However, these general-purpose models struggle to produce specialized diagrams that demand logical coherence, structural accuracy, and precise text rendering, often resulting in factual and topological errors \citep{tang2022daam}. To address this limitation, a dedicated research area has emerged focusing on the automated generation of structured visual designs from text. This field encompasses both commercial tools (e.g., DiagramGPT \citep{zala2023diagrammergpt}, Miro AI \citep{miroai2024}) that translate simple instructions into diagrams, and more advanced academic systems. While the former are ill-suited for parsing long-form documents like academic papers, the latter, such as SciDoc2Diagrammer \citep{mondal2024scidoc2diagrammer} and MagicGeo \citep{wang2025magicgeo}, employ sophisticated pipelines to generate factually-grounded diagrams from scientific literature. A critical limitation, however, is that these advanced systems typically produce static, non-interactive image outputs. Consequently, a method for automatically converting the complex content of a full academic paper into an editable and interactive diagram remains an open challenge.

\subsection{Vision-Language Agents for Document Understanding
}
With the deep integration of LLMs and visual capabilities, Vision-Language Agents have demonstrated immense potential in executing complex multimodal tasks. In the context of our task, the most relevant application is the automated generation of presentations from long-form documents.

PPTAgent \citep{pptagent2025} is a representative work in this domain. It emulates the human workflow of creating presentations by adopting a two-stage ``analyze-draft-edit" methodology. It first analyzes a reference document to learn its content layout and design patterns, then generates an outline and iteratively produces editing actions to create new slides. Such works have achieved notable progress in content summarization and initial layout generation. However, they are constrained by the limited context window of current models, which leads to performance degradation when processing exceptionally long documents. Furthermore, their primary objective is to generate a sequence of slides for presentation purposes, rather than to precisely extract and visualize the core technical model of a paper. Earlier work, such as Li et al. \citep{li2021topic}, also explored the generation of topic-aware slides from academic papers but similarly focused on content summarization rather than the construction of model diagrams.

%-------------------------------------------------------------------------
\section{Paper2SysArch Benchmark}
\subsection{Design Principles and Motivation}

A breakthrough in the automated generation of scientific diagrams necessitates an objective, comprehensive, and reproducible evaluation benchmark.

The limitations of existing models are the primary motivation for our design. Current Text-to-Image (T2I) models, such as DALL-E 3 \cite{zhou2023ethical} and Stable Diffusion \cite{stablediffusion2022}, are essentially ``black-box" pixel synthesizers. They fail to comprehend the underlying logical structure of a diagram, such as parent-child containment relationships or directed connections representing data flow. Consequently, the images they produce, while perhaps aesthetically passable, are often semantically and topologically erroneous. Furthermore, they output non-editable bitmaps, failing to meet the iterative modification needs of researchers.

The shortcomings of current agent-based approaches also inform our design. Although some agent systems (e.g., PPTAgent) can process long documents to generate presentations, their primary objective is content summarization and page layout, not the deep structural parsing of a single, complex technical model. They often fail to precisely capture and reconstruct the intricate hierarchies and interaction details within a method—the very essence of a system architecture diagram.

Drawing upon these insights, we establish the core design principle of the Paper2SysArch Benchmark: a focus on structure-centric semantics over pure visual matching. This implies that our benchmark must transcend simple pairings of ``paper PDF to example diagram." To truly assess whether a system ``understands" a paper, we require a machine-readable ground truth that describes the diagram's intrinsic structure.

To this end, our benchmark adheres to the following key design principles:

Complete Hierarchical Structure and Node Descriptions: Our ground truth is not an image but rather a standardized graphJSON format. It explicitly represents each diagram as a hierarchical graph. Within this structure, each node (representing a system component) contains detailed attributes such as text descriptions and types; edges define the directed connections (data or control flow) between components; and parent-child relationships clearly delineate the system's modularity and containment structure. This design elevates evaluation beyond the pixel level to the logical and semantic planes.

Multi-dimensional Orthogonal Evaluation: A high-quality architecture diagram must excel across multiple dimensions. Accordingly, our evaluation framework is structured into three relatively independent layers:

\begin{itemize}
    \item Semantic Layer: Evaluates the logical fidelity of the generated diagram, i.e., whether the nodes, connections, and hierarchy align with the intent of the source text.
    \item Layout Layer: Assesses the diagram's readability and clarity, identifying issues such as element overlaps and edge crossings.
    \item Visual Layer: Appraises the aesthetic quality and information efficiency of the final rendered image, including icon-semantic consistency and text clarity.
\end{itemize}
Automation and Scalability: To enable large-scale processing and ensure objectivity, the entire evaluation pipeline is designed for high automation. We leverage a suite of specialized Evaluation Agents that simulate the adjudication process of human experts, thereby facilitating rapid, consistent, and reproducible scoring of the generated diagrams.

\subsection{Ground-Truth Dataset Construction}
The cornerstone of the Paper2SysArch Benchmark is the construction of a high-quality, large-scale ground-truth dataset. This process involves the precise conversion of human-created visual knowledge, embedded within academic papers, into machine-readable structured data. The entire construction pipeline adheres to rigorous standards and relies on extensive collaboration with domain experts.

\paragraph{Data Source}
Our raw data is sourced exclusively from top-tier academic conferences in computer vision and artificial intelligence, primarily including CVPR, NeurIPS, MLSys, \dots. These sources were selected based on the following considerations: (1) these papers represent the forefront of research in their respective fields; (2) they typically contain high-quality, information-dense system architecture diagrams; and (3) their methodologies are broadly representative of the field.

\paragraph{Selection Criteria}
\textbf{Centrality and Representativeness: }The diagram must be central to the paper's methodology section, clearly illustrating the main architecture or workflow of the proposed model.
\textbf{Structural Clarity:} The diagram's layout and logic must be relatively unambiguous, avoiding overly cluttered or equivocal representations.
\textbf{Sufficient Complexity:} We excluded overly simplistic diagrams (e.g., those with fewer than three components) to ensure the dataset remains challenging.
\textbf{Self-Containment:} The diagram should be largely self-explanatory, with its primary components and processes understandable from the figure itself and its accompanying caption.

\paragraph{Expert Collaboration and Annotation Process}
The construction of the ground truth is a labor-intensive, expert-driven process. We recruited a team of PhD students with experience in scientific writing to serve as expert annotators. The annotation process proceeded as follows:
First, domain experts meticulously annotated \textbf{108 representative papers} to distill a set of core quality principles, such as semantic accuracy, structural completeness, hierarchical consistency, logical flow integrity, and appropriate granularity. Subsequently, these principles were translated into concrete instructions to guide a multi-agent collaborative system in automatically performing structured annotation on the remaining 2,900 papers. Finally, all annotated data underwent a unified cross-validation and final review process to ensure consistency and high quality across the entire dataset.

\subsection{Dataset Statistics and Analysis}
% 图表分析：节点/边/深度的分布，类别多样性等

\begin{figure}[ht]
    \centering
    \includegraphics[width=0.7\linewidth]{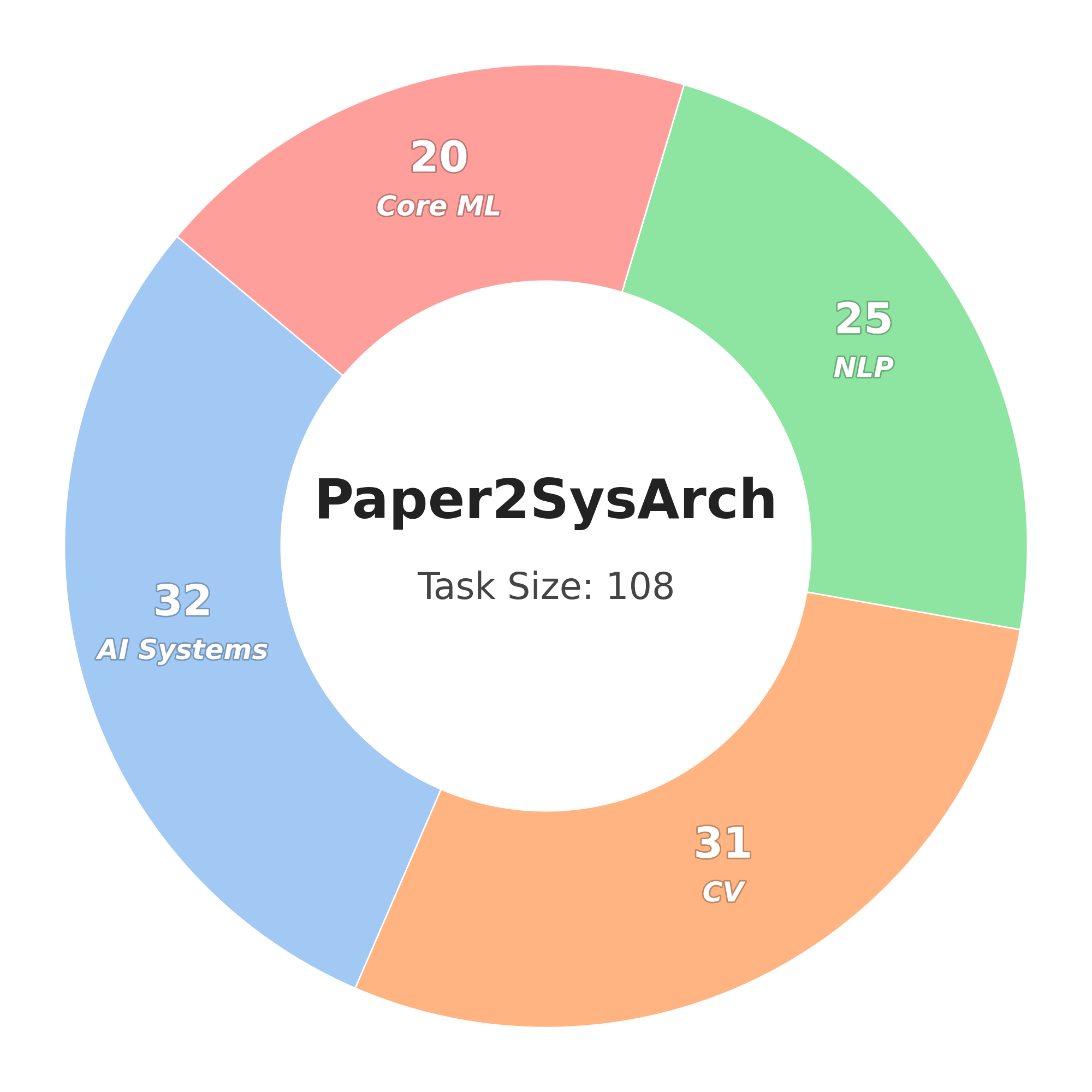}
    \caption{Domain distribution of system architecture diagrams in the 108 papers.}
    \label{fig:statistics}
\end{figure}

To validate the effectiveness of Paper2SysArch as an evaluation benchmark, we conducted a multi-dimensional statistical analysis to characterize its scale, coverage, complexity, and diversity.

\paragraph{Domain Distribution and Balance}
The dataset comprises 108 system architecture diagrams corresponding to \textbf{manually selected} papers from top-tier conferences. As shown in Figure \ref{fig:statistics}, these samples are balancedly distributed across four core domains: AI Systems (32), Computer Vision (31), Natural Language Processing (25), and Core Machine Learning (20). This balanced cross-domain coverage ensures the breadth and fairness of the benchmark.

\begin{figure*}[ht]
    \centering
    \includegraphics[width=0.8\textwidth]{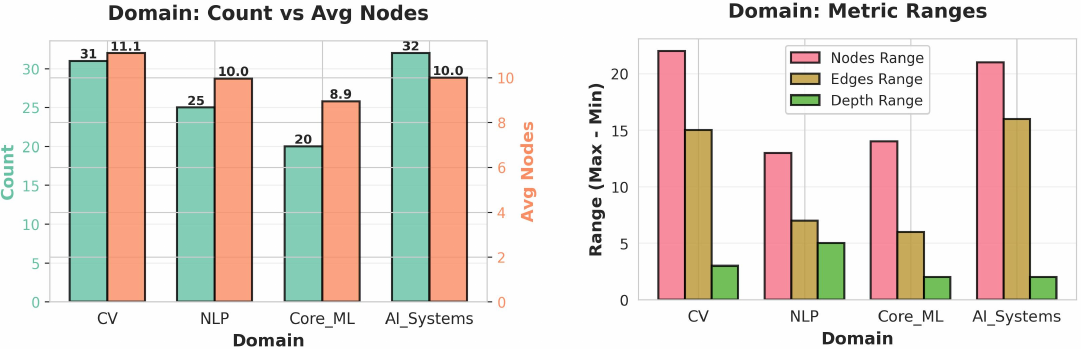}
    \caption{Structural diversity and complexity metrics across domains.}
    \label{fig:statistics-2}
\end{figure*}

\paragraph{Structural Complexity}
 The diagrams in the dataset feature significant structural complexity, avoiding simple ``toy" examples. The average number of nodes in each domain is relatively high (CV: 11.1, AI Systems/NLP: 10.0, Core ML: 8.9), which ensures that the evaluation tasks are sufficiently challenging.

\paragraph{Structural Diversity}
In addition to high average complexity, the dataset also exhibits substantial structural diversity. As detailed in Table \ref{fig:statistics-2}, each domain shows a wide range of variation in metrics such as the number of nodes, edges, and depth. For instance, the range for the number of nodes in both the CV and AI Systems domains exceeds 20. This indicates that the dataset covers a spectrum of architectures, from simple to highly complex, which is crucial for evaluating the generalization and robustness of models.

\subsection{Automated Evaluation Framework}

\subsubsection{Evaluation Agents}
\begin{figure}[ht!]
    \centering
    \includegraphics[width=1.0\linewidth]{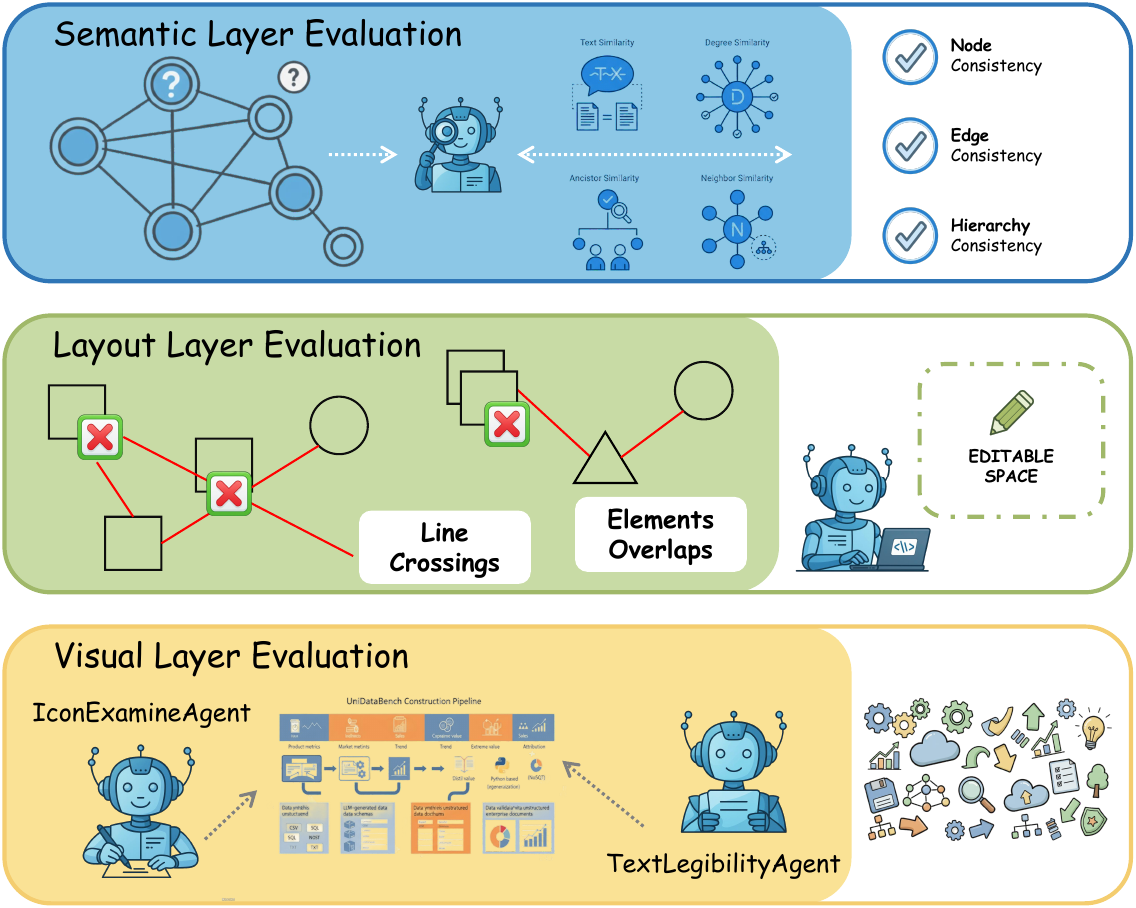}
    \caption{Overall evaluation pipeline of the Paper2SysArch benchmark.}
    \label{fig:benchmaker}
    \vspace{-10pt}
\end{figure}

As illustrated in Figure \ref{fig:benchmaker}, which outlines the construction and evaluation pipeline of our benchmark, we have designed and developed a comprehensive evaluation benchmark named Paper2SysArch to facilitate an objective, thorough, and reproducible quality assessment of automatically generated system architecture diagrams. The core objective of this benchmark is to establish a multi-dimensional quantitative evaluation framework. This framework systematically measures the consistency between a generated diagram and a GT diagram—manually constructed by domain experts—across three orthogonal dimensions: Semantic Accuracy, Layout Rationality, and Visual Expressiveness. The construction and execution of this benchmark are heavily reliant on a collaborative suite of specialized agents that programmatically simulate the perceptual, comprehension, and judgmental capabilities of human experts, thereby enabling the automation and scalability of the evaluation process.

First, to construct the GT dataset required for evaluation, the \textit{GraphExtractAgent} plays a pivotal role. This agent is responsible for extracting structural information from diverse input sources (such as diagram images and non-standard format description files) and precisely converting it into our defined, standardized hierarchical graphJSON format. 

Second, within the evaluation pipeline, the \textit{SystemUnderstandAgent} serves as a core semantic understanding engine. During the visual-layer assessment, it is tasked with deeply analyzing the visual content of both the generated diagram and the GT diagram to produce respective summary descriptions of the system architecture. By comparing the semantic similarity of these descriptions, we can quantitatively evaluate the performance of the generated diagram in terms of information fidelity.

Finally, a series of review agents based on VLMs \cite{wang2025internvl3_5,li2024llava,hurst2024gpt} constitutes the executive core of the automated evaluation. The \textit{IconExaminAgent} is responsible for examining the semantic relevance of icons, determining whether they appropriately convey the function of their corresponding components. The \textit{LayoutExamineAgent} systematically detects layout flaws within the diagram, such as line crossings and element overlaps. Meanwhile, the \textit{TextLegibilityAgent} focuses on assessing the readability of textual elements, identifying issues like blurry or distorted text. Collectively, these VLM-based agents provide comprehensive and objective quantitative scores for the generated architecture diagrams across the layout and visual dimensions.

\subsubsection{Semantic Layer Evaluation}

The semantic-level evaluation aims to quantify the extent to which a generated graph faithfully reproduces the system architecture articulated in the original paper. This process encompasses the accurate representation of system components, their hierarchical relationships, and the data or control flows between them. The evaluation methodology consists of two core steps: a node matching algorithm and the measurement of multi-dimensional semantic consistency.

\paragraph{Node Matching Algorithm}
Before comparing the structure and content of the two graphs, it is imperative to establish a reliable correspondence between the node set of the generated graph, $N_g$, and that of the ground-truth graph, $N_{gt}$. To this end, we devise a two-stage iterative matching algorithm. The core of this algorithm is the computation of a composite similarity score, $S(n_g, n_{gt})$, for each candidate node pair $(n_g, n_{gt})$, where $n_g \in N_g$ and $n_{gt} \in N_{gt}$. This score is a weighted sum of several heterogeneous features:
\begin{multline}
\label{eq:similarity_score} 
S(n_g, n_{gt}) = w_t S_{\text{text}} + w_d S_{\text{degree}} \\
+ w_a S_{\text{ancestor}} + w_n S_{\text{neighbor}}
\end{multline}
where $w_t, w_d, w_a,$ and $w_n$ are the respective weights for text, degree, ancestor, and neighbor similarities. These similarity components are calculated as follows:

First, \textbf{Text Similarity} ($S_{\text{text}}$) leverages a pre-trained language model (e.g., BERT \cite{he2021deberta,devlin2019bert}) to compute the cosine similarity between the semantic vectors of the two nodes' names and their content descriptions, thereby capturing the congruence of their intrinsic meanings.

Second, \textbf{Degree Similarity} ($S_{\text{degree}}$) measures the structural connectivity of the nodes. It is defined as:
\begin{multline}
S_{\text{degree}}(n_g, n_{gt}) = \exp\big(-\big(|\text{deg}^+(n_g)
- \text{deg}^+(n_{gt})| \\
{}+ |\text{deg}^-(n_g) - \text{deg}^-(n_{gt})|\big)\big)
\end{multline}

where $\text{deg}^+$ and $\text{deg}^-$ denote the out-degree and in-degree of a node, respectively.

Third, \textbf{Ancestor Similarity} ($S_{\text{ancestor}}$) assesses whether the nodes occupy consistent macroscopic positions within their respective hierarchical structures. This is achieved by recursively summing the text similarities of their parent, grandparent, and so on, up to the root nodes.

Finally, \textbf{Neighbor Similarity} ($S_{\text{neighbor}}$) is calculated as the proportion of a node's neighbors that have already been matched. This component leverages locally established structural information to infer further matches.

The matching process is conducted in two rounds. The first round assigns a significantly high weight to text similarity ($w_t$) to confidently match nodes with distinct and unambiguous textual descriptions. After the first round, the second round reduces the weight $w_t$ while increasing the weights for ancestor and neighbor similarities ($w_a$ and $w_n$). This strategy utilizes the structural anchors established in the first round to resolve matches for nodes with ambiguous or vague textual information.

Once the correspondence between nodes is established, we quantify semantic consistency across three dimensions: nodes, edges, and hierarchy. The detailed definitions of the metrics used for evaluation are provided in supplementary material.
% Appendix \ref{appendix:A}.

\subsubsection{Layout Layer Evaluation}
The layout layer evaluation aims to quantify the layout quality and visual readability of the generated graph, which directly impact the user's information processing efficiency. We leverage the aforementioned VLM-powered agents (specifically, the \textit{LayoutExamineAgent}) to programmatically detect common layout defects. The primary defects detected are edge crossings, element overlaps, and text overflows, with their counts denoted as $C_{\text{cross}}, C_{\text{overlap}},$ and $C_{\text{overflow}}$, respectively. The final layout score, $S_{\text{layout}}$, is calculated using a penalty-based mechanism. It starts at a maximum of 1.0 and deducts a fixed penalty $\delta$ (e.g., 0.1) for each detected defect, down to a minimum of zero. The formula :
\begin{multline}
    S_{\text{layout}} = \max\big(0,\, 1.0 - \delta \cdot (C_{\text{cross}} \\
    {}+ C_{\text{overlap}} + C_{\text{overflow}})\big)
\end{multline}

Additionally, we qualitatively assess the layout's modifiability by examining whether sufficient white space is preserved for potential manual adjustments by the user.

\subsubsection{Visual Layer Evaluation}
The visual layer evaluation is designed to measure the aesthetic quality of the image and the efficiency of semantic conveyance through visual elements, with these assessments being performed automatically by designated agents. For icon relevance, the \textit{IconExamineAgent} utilizes a VLM to generate a textual description for each auto-generated icon and then calculates the semantic similarity score between this description and the name or functional description of the icon's parent module; this score reflects the extent to which the icon aids in semantic understanding. To assess overall comprehension difficulty, the \textit{SystemUnderstandAgent} evaluates information fidelity by generating summary descriptions of the entire system based on the GT graph and the generated graph separately, and then computes the semantic similarity between the two summaries. A higher similarity indicates that the generated graph's ability to convey core system information is closer to the ground truth. Finally, for text legibility, the \textit{TextLegibilityAgent} is responsible for detecting blurry or illegible text caused by issues such as low resolution, artifacts, or distortion. Its scoring mechanism is similar to the layout evaluation, starting from a perfect score and deducting points for each instance of unclear text found.

%-------------------------------------------------------------------------
\section{Paper2SysArch Agent}
\label{sec:agent}
% \textcolor{red}{Mock!}
\begin{figure}[ht!]
    \centering
    \includegraphics[width=1.0\linewidth]{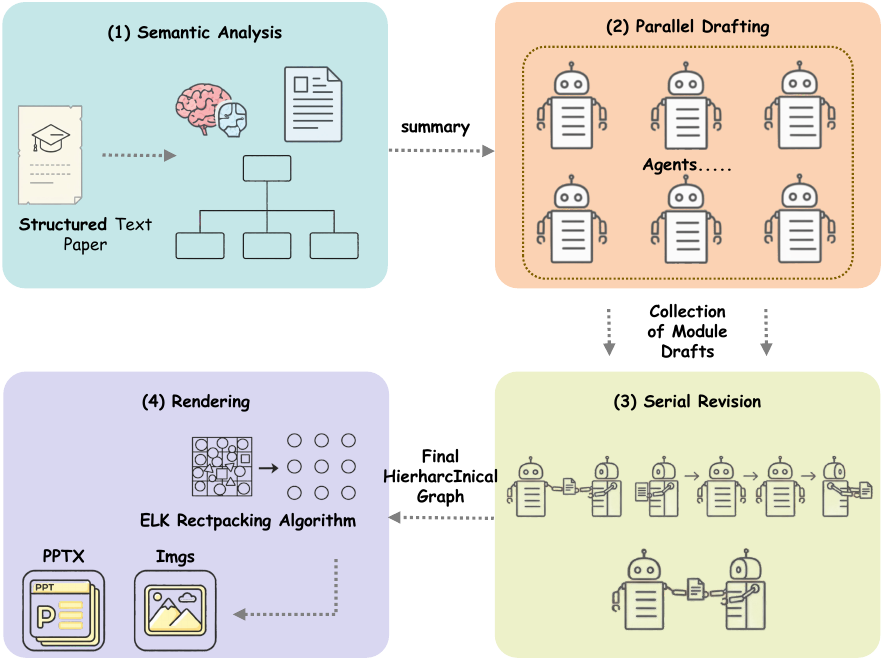}
\caption{Overview of the Paper2SysArch Agent.}
    \label{fig:overview}
\end{figure}

Translating complex system descriptions from academic papers into clear architectural diagrams is a significant challenge. We propose the Paper2SysArch Agent to address this. It is an intelligent system designed to automate this process. The system's core contribution is its novel generation pipeline. This pipeline uses a custom hierarchical graph representation. It systematically maps unstructured text to formal architectural diagrams. The overall system architecture is illustrated in Figure~\ref{fig:overview}.

\subsection{Hierarchical Graph Representation}
\label{ssec:representation}

Traditional graph structures often fail to describe complex systems clearly. They lack hierarchical constraints, leading to confusing ``spaghetti" connections. To solve this problem, we propose a novel hierarchical graph representation. This method is not a flat collection of nodes and edges. Instead, it is built on a strict, three-layer hierarchy:

\begin{enumerate}
    \item \textbf{Module Layer:} The highest level of abstraction. It defines the system's main functional components or processing stages.
    \item \textbf{Tool/Data Layer:} This layer populates modules with concrete items. These include execution units like models and algorithms, as well as data objects.
    \item \textbf{Component Layer:} The lowest level. It provides semantic details for tools and data, such as names, descriptions, or icons.
\end{enumerate}

The key innovation is a strict rule for connections: directed edges are only allowed between sibling nodes under the same parent. This principle forces cross-module interactions to be represented at a higher level. This ensures the graph structure remains clear and logical. It prevents structural ambiguity by design.

\subsection{Automated Generation Pipeline}
We designed a multi-stage pipeline to automatically convert text into our hierarchical graph representation. The pipeline breaks down graph construction into several steps, moving from abstract concepts to concrete details.

The process starts with Semantic Analysis and Decomposition. A LLM analyzes the paper to identify core functional modules. This builds the top-level framework of the graph. Next, the system uses a Distributed Drafting and Centralized Revision strategy. A multi-agent system drafts the details for each module in parallel. This quickly generates a preliminary architecture. A serial revision phase then integrates all drafts and checks for consistency. This step ensures unique IDs, uniform terminology, and valid connections. The final stage is Layout and Rendering. An automatic layout algorithm arranges all nodes and edges for optimal readability. The system also programmatically generates visual assets like icons. Finally, it renders a professional, presentation-ready architecture diagram. More details in supplementary material.

% \begin{figure}
%     \centering
%     \includegraphics[width=1\linewidth]{agent.png}
%     \caption{Enter Caption}
%     \label{fig:placeholder}
% \end{figure}

%-------------------------------------------------------------------------
\section{Experiments}
\label{sec:experiments}
% The content for your experiments goes here.
% This section corresponds to "5 Experiments" from your image.
\subsection{Experimental Setup}
To comprehensively evaluate the performance of our system, this section details the dataset, evaluation metrics, baselines, and implementation specifics used in our experiments.

\paragraph{Dataset}
All experiments are conducted on an expertly curated subset of 108 papers from the \texttt{Paper2SysArch Benchmark} to ensure a gold-standard evaluation. The full dataset, comprising 3000 pairs, is publicly available to facilitate further research by the community.

\paragraph{Evaluation Metrics}
We employ a three-tiered evaluation framework for quantitative assessment, encompassing Semantic, Layout, and Visual aspects. The final Overall Score is computed as a weighted average, with a higher weight assigned to the semantic component to underscore the importance of accurate content understanding. The formula is defined as:
\begin{equation}
\label{eq:overall_score}
\text{Overall} = 0.3 \times \text{Semantic} + 0.3 \times \text{Layout} + 0.4 \times \text{Visual}
\end{equation}

\paragraph{Baselines}
We compare our system against two categories of baselines:
\begin{itemize}
    \item \textbf{Text-to-Image Baselines:} This category includes \textbf{DALL-E 3} and \textbf{Nanobanana}. These baselines primarily relies on Diffusion models to generate pixel-based images in an end-to-end fashion. However, they may suffer from instability in generation and a lack of post-hoc editability.
    
    \item \textbf{Code-to-Image Baselines:} This category includes \textbf{GPT-4o + GraphViz}. The baseline operates via a two-stage pipeline: first, a Large Language Model generates a structured, code-based representation; second, this representation is executed by a rendering engine to synthesize the image.
\end{itemize}
For a fair comparison, we construct detailed prompts using each paper's title, abstract, and core methodology sections. We then select the best-generated output from each model for evaluation.

\paragraph{Implementation Details}
\texttt{Paper2SysArch} agent leverages \texttt{GPT-4o} \cite{hurst2024gpt} as its core, functioning as both the LLM for textual understanding and the VLM for multimodal tasks. For the node matching component, semantic similarity scores are computed using a \texttt{Sentence-BERT} \cite{reimers2019sentence}.

\subsection{Main Performance}
\begin{table}[!h]
  \centering
  \caption{\textbf{Performance Comparison on System Architecture Diagram Generation.}}
  \label{tab:comparison}

  \footnotesize
  \setlength{\tabcolsep}{4pt}

  \begin{adjustbox}{width=\columnwidth,center}
    \begin{tabular}{@{}lccccc@{}}
      \toprule
      \textbf{Method} & \textbf{Semantic} & \textbf{Layout} & \textbf{Visual} & \textbf{Overall} & \textbf{Editable} \\
      \midrule
      \multicolumn{6}{l}{\textit{Baseline: Text‐to‐Image Generation}} \\
      DALL‐E 3       & 29.9 & 20.7 & 65.2 & 41.3 & \textcolor{red}{\ding{55}} \\
      Nanobanana     & 31.2 & 80.3 & 72.8 & 62.6 & \textcolor{red}{\ding{55}} \\
      \midrule
      \multicolumn{6}{l}{\textit{Baseline: Code‐to‐Image Generation}} \\
      GPT‐4o + GraphViz & 42.0 & 85.5 & 71.2 & 66.7 & \textcolor{green!60!black}{\ding{51}} \\
      \midrule
      \multicolumn{6}{l}{\textit{Our Agent‐based Approach}} \\
      \rowcolor{green!10}
      Paper2SysArch (Qwen2.5‐VL‐72B) & 38.7 & 76.8 & 84.6 & 68.5 & \textcolor{green!60!black}{\ding{51}} \\
      \rowcolor{green!10}
      Paper2SysArch (GPT‐4o)          & 29.8 & 83.9 & 87.3 & 69.0 & \textcolor{green!60!black}{\ding{51}} \\
      \bottomrule
    \end{tabular}
  \end{adjustbox}
\end{table}

\begin{table*}[!t]
\centering
\caption{Comparison of different Benchmarks.}
\label{tab:benchmark_comparison}

% 设定列格式：
% l: 第一列 Benchmark 左对齐
% cccc: 中间4列数据居中对齐
% l: 最后一列 Eval Dims 左对齐（因为维度描述通常文字较长，左对齐更易读，若内容短可改为 c）
\begin{tabular}{l ccccc}
\toprule

% --- 表头第一行 (The Macro Layer) ---
% Col 1: Benchmark (跨2行)
\multirow{2}{*}{Benchmarks} 
% Col 2-5: Dataset (跨4列，居中)
& \multicolumn{4}{c}{Dataset} 
% Col 6: Evaluation Dimensions (跨2行)
& \multirow{2}{*}{Evaluation Dimensions} \\

% --- 装饰线 ---
% 只在 Dataset 下面画线，左右留白 (lr)，从第2列画到第5列
\cmidrule(lr){2-5} 

% --- 表头第二行 (The Micro Layer) ---
% Col 1: 留空 (被上面的 Benchmark 占了)
& Papers & Conferences & Domains & Data Modals 
% Col 6: 留空 (被上面的 Eval Dims 占了)
& \\

\midrule

% --- 数据示例 ---
Paper2Poster \cite{pang2025paper2poster} & 100 & 3 & 3 & Papers + Posters & 10 \\
Paper2Video \cite{zhu2025paper2video}  & 101 & 9 & 3 & Papers + Videos & 3 \\
DiagramEval \cite{liang2024diagrameval}  & \multicolumn{4}{c}{No Dataset, Only a Collection Pipeline} & 2\\
\rowcolor{green!10}
\textbf{Paper2SysArch (Ours)}  & 3000 & 8 & 4 & Papers + SysArchFigures & 7 \\

\bottomrule
\end{tabular}
\end{table*}

As shown in Table \ref{tab:comparison}, our Paper2SysArch (GPT-4o) method achieves an overall score of 69.0, surpassing all automated baselines, including traditional text-to-image models like DALL-E 3 (41.3) and code-generation solutions such as GPT-4o + GraphViz (66.7). This lead is primarily attributed to its exceptional performance in the Visual (87.3) and Layout (83.9) dimensions, demonstrating the effectiveness of our agent-based approach in generating high-quality, well-structured diagrams. However, all automated methods scored relatively low on semantic understanding, and our method (29.8) also failed to outperform the code-based baseline (42.0) in this regard. This highlights that accurately capturing and reproducing the core logical structure of the paper remains a key challenge.

\subsection{Comparison with Other Benchmarks}
As shown in Table \ref{tab:benchmark_comparison}, Paper2SysArch significantly outperforms existing benchmarks in terms of data scale, diversity, and evaluation depth. Our dataset comprises \textbf{3,000} papers from \textbf{8} top-tier conferences across \textbf{4} research fields, far exceeding other works in scope and enabling tasks of greater complexity. Furthermore, unlike benchmarks that focus on poster or video generation, Paper2SysArch is specifically centered on the task of generating system architecture diagrams—a research direction that has previously lacked systematic investigation. We have also constructed a more comprehensive evaluation framework with \textbf{7} dimensions, surpassing the assessment depth of existing benchmarks.

\subsection{Comparative Analysis with Baselines
}
% \subsection{Ablation Study}
% 完整系统 (Full System): 即 Paper2SysArch。
% 无顶层设计 (w/o Top-level Design): 在语义提取后，直接让模块细化阶段处理所有概念，不进行宏观布局规划。
% 无模块细化 (w/o Module Refinement): 在顶层设计后，直接进行图形渲染，跳过对模块内部结构和连接的详细设计。

\subsection{Case Study}

\begin{figure}[htbp]
    \centering
    \includegraphics[width=1\linewidth]{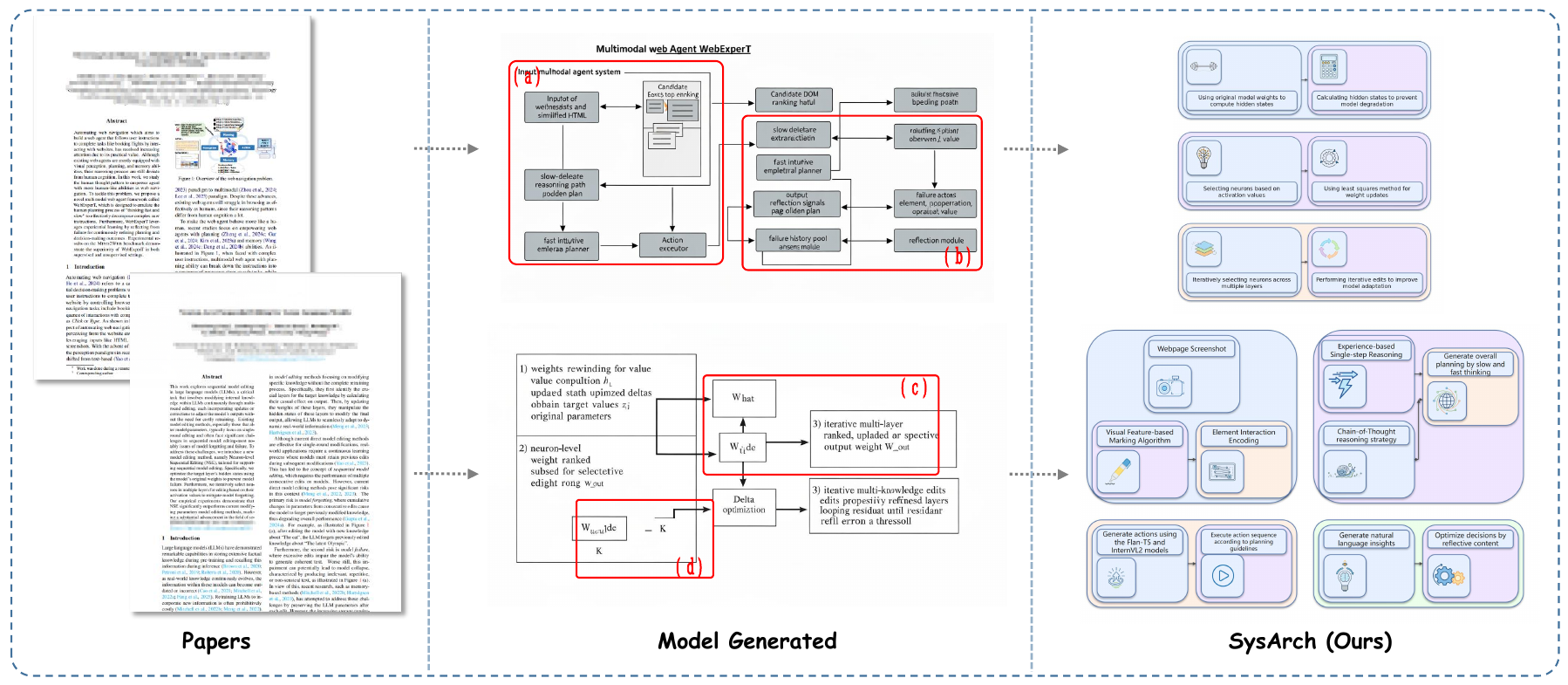}
    \caption{Case Study: Comparison Between SysArch and Baseline Model.}
    \label{fig:casestudy}
    \vspace{-5pt}
\end{figure}

To visually demonstrate and evaluate the performance of our proposed SysArch method in generating system architecture diagrams, we conducted a case study. For this study, we selected two representative academic papers and present a qualitative comparison between the diagrams generated by DALL-E 3 and those produced by our method.

As can be clearly observed from the Figure \ref{fig:casestudy}, the diagrams generated by the baseline model exhibit significant shortcomings in several aspects. While it attempts to capture some key components, it performs poorly in representing inter-component connections, ensuring a coherent overall layout, and maintaining the accuracy of text labels. For instance, the baseline's outputs include: (a) unspecified visual artifacts; (b) a high prevalence of garbled text and disorganized connections; (c) an extraneous module labeled ``what"; and (d) other randomly generated elements. These errors lead to a significant semantic deviation from the original design, resulting in poor readability and a failure to accurately convey the paper's core architectural concepts. Furthermore, even on the occasions when the generative model produces a reasonably structured diagram, its output is not amenable to direct secondary editing.

In contrast, the diagrams generated by our method align closely with the original diagrams from the papers in terms of visual fidelity, structural layout, and semantic logic. Our method not only accurately identifies and renders all key components but also precisely reproduces their hierarchical relationships and dependency paths. Moreover, our method can directly output the diagrams as editable PPTX files, a key advantage for practical use.

%-------------------------------------------------------------------------
\section{Conclusions}
\label{sec:conclusions}
In conclusion, we present Paper2SysArch, the first large-scale benchmark dedicated to generating system architecture diagrams from scientific papers. It provides a larger and more diverse dataset compared to prior work and introduces a more comprehensive evaluation methodology. We hope this benchmark will serve as a useful tool and testbed for future research and model development in this field.

Building upon this benchmark, we also developed the Paper2SysArch agent, capable of generating semantically coherent system architecture diagrams in an automated and controllable manner. This system provides a solid foundation for future research in this domain and demonstrates a promising path forward for the task.

However, our agent's reliance on pre-defined open-source algorithms for layout generation can lead to inflexibility. Furthermore, a considerable gap in aesthetic quality remains when compared to human-authored diagrams. Both of these areas represent key avenues for future research.

%-------------------------------------------------------------------------
% \section*{Acknowledgements}
% The content for your acknowledgements goes here. This section is unnumbered.
% This section corresponds to "7 Acknowledgements." from your image.

\clearpage
\newpage
%%%%%%%%% REFERENCES
{
    \small
    \bibliographystyle{ieeenat_fullname}
    \bibliography{main}

\begin{thebibliography}{32}
\providecommand{\natexlab}[1]{#1}
\providecommand{\url}[1]{\texttt{#1}}
\expandafter\ifx\csname urlstyle\endcsname\relax
  \providecommand{\doi}[1]{doi: #1}\else
  \providecommand{\doi}{doi: \begingroup \urlstyle{rm}\Url}\fi

\bibitem[dia(2024)]{diagramgenbenchmark2024}
Diagramgenbenchmark: A benchmark dataset for text-to-diagram generation.
\newblock
  \url{https://huggingface.co/datasets/DiagramAgent/DiagramGenBenchmark}, 2024.
\newblock Accessed: 2024-11-12.

\bibitem[Chen et~al.(2024{\natexlab{a}})Chen, Liu, Chen, Shi, and
  Zhao]{chen2024textdiffuser}
Jingye Chen, Yupan Liu, Hao Chen, Jianqi Shi, and Zhouhui Zhao.
\newblock Textdiffuser-2: Unleashing the power of language models for text
  rendering.
\newblock In \emph{European Conference on Computer Vision}, pages 392--409.
  Springer, 2024{\natexlab{a}}.

\bibitem[Chen et~al.(2024{\natexlab{b}})Chen, Mei, Fan, and
  Wang]{chen2024opportunities}
Minshuo Chen, Song Mei, Jianqing Fan, and Mengdi Wang.
\newblock Opportunities and challenges of diffusion models for generative ai.
\newblock \emph{National Science Review}, 11\penalty0 (12):\penalty0 nwae348,
  2024{\natexlab{b}}.

\bibitem[Chen et~al.(2024{\natexlab{c}})Chen, Zhang, Xu, Ren, and
  Yang]{chen2024viseval}
Nan Chen, Yuge Zhang, Jiahang Xu, Kan Ren, and Yuqing Yang.
\newblock Viseval: A benchmark for data visualization in the era of large
  language models.
\newblock \emph{IEEE Transactions on Visualization and Computer Graphics},
  2024{\natexlab{c}}.

\bibitem[Devlin et~al.(2019)Devlin, Chang, Lee, and Toutanova]{devlin2019bert}
Jacob Devlin, Ming-Wei Chang, Kenton Lee, and Kristina Toutanova.
\newblock Bert: Pre-training of deep bidirectional transformers for language
  understanding.
\newblock In \emph{Proceedings of the 2019 Conference of the North American
  Chapter of the Association for Computational Linguistics: Human Language
  Technologies, Volume 1 (Long and Short Papers)}, pages 4171--4186,
  Minneapolis, Minnesota, 2019. Association for Computational Linguistics.

\bibitem[He et~al.(2021)He, Liu, Gao, and Chen]{he2021deberta}
Pengcheng He, Xiaodong Liu, Jianfeng Gao, and Weizhu Chen.
\newblock Deberta: Decoding-enhanced bert with disentangled attention.
\newblock In \emph{Proceedings of the 9th International Conference on Learning
  Representations (ICLR)}, 2021.

\bibitem[Hong et~al.(2023)Hong, Zheng, Chen, Cheng, Zhang, Wang, Wang, Yau,
  Lin, Zhou, et~al.]{hong2023chatdev}
Sirui Hong, Xiawu Zheng, Jonathan Chen, Yuheng Cheng, Jinlin Zhang, Ceyao Wang,
  Zili Wang, Steven Ka~Shing Yau, Zijuan Lin, Liyang Zhou, et~al.
\newblock Chatdev: Communicative agents for software development.
\newblock \emph{arXiv preprint arXiv:2307.07924}, 2023.

\bibitem[Hurst et~al.(2024)Hurst, Lerer, Goucher, Perelman, Ramesh, Clark,
  Ostrow, Welihinda, Hayes, Radford, et~al.]{hurst2024gpt}
Aaron Hurst, Adam Lerer, Adam~P Goucher, Adam Perelman, Aditya Ramesh, Aidan
  Clark, AJ Ostrow, Akila Welihinda, Alan Hayes, Alec Radford, et~al.
\newblock Gpt-4o system card.
\newblock \emph{arXiv preprint arXiv:2410.21276}, 2024.

\bibitem[Li et~al.(2024)Li, Zhang, Zhang, Zhang, Li, Li, Ma, and
  Li]{li2024llava}
Feng Li, Renrui Zhang, Hao Zhang, Yuanhan Zhang, Bo Li, Wei Li, Zejun Ma, and
  Chunyuan Li.
\newblock Llava-next-interleave: Tackling multi-image, video, and 3d in large
  multimodal models.
\newblock \emph{arXiv preprint arXiv:2407.07895}, 2024.

\bibitem[Li et~al.(2021)]{li2021topic}
Yew~Lin Li et~al.
\newblock Towards topic-aware slide generation from academic papers.
\newblock \emph{Conference proceedings}, 2021.

\bibitem[Liang et~al.(2024)]{liang2024diagrameval}
Chumeng Liang et~al.
\newblock Diagrameval: Evaluating llm-generated diagrams via graphs.
\newblock \emph{arXiv preprint arXiv:2510.25761}, 2024.

\bibitem[Liao et~al.(2025)Liao, Dai, Wang, Yang, Liu, Barker, Rintamaki,
  Shoeybi, Catanzaro, and Ping]{liao2025challenges}
Wenliang Liao, Nayeon Dai, Boxin Wang, Zhuolin Yang, Zihan Liu, Jon Barker,
  Tuomas Rintamaki, Mohammad Shoeybi, Bryan Catanzaro, and Wei Ping.
\newblock Challenges in generating accurate text in images: A benchmark for
  text-to-image models on specialized content.
\newblock \emph{Applied Sciences}, 15\penalty0 (5):\penalty0 2274, 2025.

\bibitem[{Miro}(2024)]{miroai2024}
{Miro}.
\newblock Miro ai -- ai innovation workspace.
\newblock \url{https://miro.com/}, 2024.
\newblock AI-powered collaborative whiteboard platform.

\bibitem[Mondal et~al.(2024)Mondal, Li, Hou, Natarajan, Garimella, and
  Boyd-Graber]{mondal2024scidoc2diagrammer}
Ishani Mondal, Zongxia Li, Yufang Hou, Anandhavelu Natarajan, Aparna Garimella,
  and Jordan~Lee Boyd-Graber.
\newblock Scidoc2diagrammer-maf: Towards generation of scientific diagrams from
  documents guided by multi-aspect feedback refinement.
\newblock In \emph{Findings of the Association for Computational Linguistics:
  EMNLP 2024}, pages 13342--13375, Miami, Florida, USA, 2024. Association for
  Computational Linguistics.

\bibitem[Nichol and Dhariwal(2021)]{nichol2021improved}
Alexander~Quinn Nichol and Prafulla Dhariwal.
\newblock Improved denoising diffusion probabilistic models.
\newblock In \emph{International Conference on Machine Learning}, pages
  8162--8171, 2021.

\bibitem[Pang et~al.(2025)Pang, Lin, Jian, He, and Torr]{pang2025paper2poster}
Wei Pang, Kevin~Qinghong Lin, Xiangru Jian, Xi He, and Philip Torr.
\newblock Paper2poster: Towards multimodal poster automation from scientific
  papers.
\newblock \emph{arXiv preprint arXiv:2505.21497}, 2025.

\bibitem[Podell et~al.(2023)Podell, English, Lacey, Blattmann, Dockhorn,
  Müller, Penna, and Rombach]{podell2023sdxl}
Dustin Podell, Zion English, Kyle Lacey, Andreas Blattmann, Tim Dockhorn, Jonas
  Müller, Joe Penna, and Robin Rombach.
\newblock Sdxl: Improving latent diffusion models for high-resolution image
  synthesis.
\newblock \emph{arXiv preprint arXiv:2307.01952}, 2023.

\bibitem[Reimers and Gurevych(2019)]{reimers2019sentence}
Nils Reimers and Iryna Gurevych.
\newblock Sentence-bert: Sentence embeddings using siamese bert-networks.
\newblock \emph{arXiv preprint arXiv:1908.10084}, 2019.

\bibitem[Rombach et~al.(2022{\natexlab{a}})Rombach, Blattmann, Lorenz, Esser,
  and Ommer]{rombach2022high}
Robin Rombach, Andreas Blattmann, Dominik Lorenz, Patrick Esser, and Björn
  Ommer.
\newblock High-resolution image synthesis with latent diffusion models.
\newblock In \emph{Proceedings of the IEEE/CVF Conference on Computer Vision
  and Pattern Recognition}, pages 10684--10695, 2022{\natexlab{a}}.

\bibitem[Rombach et~al.(2022{\natexlab{b}})Rombach, Blattmann, Lorenz, Esser,
  and Ommer]{stablediffusion2022}
Robin Rombach, Andreas Blattmann, Dominik Lorenz, Patrick Esser, and Bjorn
  Ommer.
\newblock High-resolution image synthesis with latent diffusion models.
\newblock \emph{CVPR}, 2022{\natexlab{b}}.

\bibitem[Song et~al.(2021)Song, Meng, and Ermon]{song2020denoising}
Jiaming Song, Chenlin Meng, and Stefano Ermon.
\newblock Denoising diffusion implicit models.
\newblock In \emph{International Conference on Learning Representations}, 2021.

\bibitem[Tang et~al.(2022)Tang, Liu, Pandey, Jiang, Yang, Kumar, Stenetorp,
  Lin, and Ture]{tang2022daam}
Raphael Tang, Linqing Liu, Akshat Pandey, Zhiying Jiang, Gefei Yang, Karun
  Kumar, Pontus Stenetorp, Jimmy Lin, and Ferhan Ture.
\newblock What the daam: Interpreting stable diffusion using cross attention.
\newblock \emph{arXiv preprint arXiv:2210.04885}, 2022.

\bibitem[Team(2024)]{gemini}
Gemini Team.
\newblock Gemini: A family of highly capable multimodal models, 2024.

\bibitem[Wang et~al.(2025{\natexlab{a}})Wang, Zhang, Yu, Wang, and
  Huang]{wang2025magicgeo}
Junxiao Wang, Ting Zhang, Heng Yu, Jingdong Wang, and Hua Huang.
\newblock Magicgeo: Training-free text-guided geometric diagram generation.
\newblock \emph{arXiv preprint arXiv:2502.13855}, 2025{\natexlab{a}}.

\bibitem[Wang et~al.(2024)]{magicgeobench2024}
Junxiao Wang et~al.
\newblock Magicgeo: Training-free text-guided geometric diagram generation.
\newblock 2024.
\newblock Benchmark dataset of 220 geometric diagram descriptions.

\bibitem[Wang et~al.(2025{\natexlab{b}})Wang, Gao, Gu, Pu, Cui, Wei, Liu, Jing,
  Ye, Shao, et~al.]{wang2025internvl3_5}
Weiyun Wang, Zhangwei Gao, Lixin Gu, Hengjun Pu, Long Cui, Xingguang Wei,
  Zhaoyang Liu, Linglin Jing, Shenglong Ye, Jie Shao, et~al.
\newblock Internvl3.5: Advancing open-source multimodal models in versatility,
  reasoning, and efficiency.
\newblock \emph{arXiv preprint arXiv:2508.18265}, 2025{\natexlab{b}}.

\bibitem[Zala et~al.(2023)Zala, Lin, Cho, and Bansal]{zala2023diagrammergpt}
Abhay Zala, Han Lin, Jaemin Cho, and Mohit Bansal.
\newblock Diagrammergpt: Generating open-domain, open-platform diagrams via llm
  planning.
\newblock \emph{arXiv preprint arXiv:2310.12128}, 2023.

\bibitem[Zala et~al.(2024{\natexlab{a}})Zala, Lin, Cho, and
  Bansal]{ai2dcaption2024}
Abhay Zala, Han Lin, Jaemin Cho, and Mohit Bansal.
\newblock Diagrammergpt: Generating open-domain, open-platform diagrams via llm
  planning.
\newblock In \emph{COLM}, 2024{\natexlab{a}}.

\bibitem[Zala et~al.(2024{\natexlab{b}})Zala, Lin, Cho, and
  Bansal]{diagrammergpt2024}
Abhay Zala, Han Lin, Jaemin Cho, and Mohit Bansal.
\newblock Diagrammergpt: Generating open-domain, open-platform diagrams via llm
  planning.
\newblock \emph{arXiv preprint arXiv:2310.12128}, 2024{\natexlab{b}}.

\bibitem[Zheng et~al.(2025)Zheng, Guan, Kong, Zheng, Zhou, Lin, Lu, He, Han,
  and Sun]{pptagent2025}
Hao Zheng, Xinyan Guan, Hao Kong, Jia Zheng, Weixiang Zhou, Hongyu Lin, Yaojie
  Lu, Ben He, Xianpei Han, and Le Sun.
\newblock Pptagent: Generating and evaluating presentations beyond
  text-to-slides.
\newblock \emph{arXiv preprint arXiv:2501.03936}, 2025.

\bibitem[Zhou and Nabus(2023)]{zhou2023ethical}
Kai-Qing Zhou and Hatem Nabus.
\newblock The ethical implications of dall-e: Opportunities and challenges.
\newblock \emph{Mesopotamian Journal of Computer Science}, 2023:\penalty0
  16--21, 2023.

\bibitem[Zhu et~al.(2025)Zhu, Lin, and Shou]{zhu2025paper2video}
Zeyu Zhu, Kevin~Qinghong Lin, and Mike~Zheng Shou.
\newblock Paper2video: Automatic video generation from scientific papers.
\newblock \emph{arXiv preprint arXiv:2510.05096}, 2025.

\end{thebibliography}
}

% The command for supplementary material has been removed as per your request.
\appendix

% 重新定义章节编号显示格式：S + 大写字母/数字
\renewcommand{\thesection}{S\arabic{section}} 
\renewcommand{\thesubsection}{S\arabic{section}.\arabic{subsection}}

\clearpage
\setcounter{page}{1}
\maketitlesupplementary

\section{Detailed Workflow and Agent Prompts}
\subsection{P2SA Agent}
Complementing the conceptual overview provided in the main paper, this section details the \textbf{formal definitions} and \textbf{implementation specifics} of the Paper2SysArch Agent. 
Specifically, we first provide the rigorous data schema and topological constraints for the Hierarchical Graph Representation. 
Subsequently, we expand the Automated Generation Pipeline into a granular, seven-step \textbf{Neuro-Symbolic} workflow, elaborating on the specific mechanisms for distributed drafting, topological regularization, and content-aware layout optimization.
\subsubsection{Hierarchical Graph Representation of SysArch}
\label{sec:graph_representation}

To structurally represent the system architecture(which is essential for our Structure-Constrained generation), we define a \textbf{Three-level Hierarchical Graph}, abstracting the system from macroscopic modules to microscopic components. Here is an example in Listing \ref{lst:json_example}.

\vspace{1mm}
\noindent \textbf{1. Hierarchical Levels.} The graph consists of three distinct layers:
\begin{itemize}
    \item \textbf{Module Level ($L_1$):} Represents the top-level abstraction of the system, corresponding to major processing phases such as Intent Recognition, Data Preprocessing, or Model Inference.
    \item \textbf{Entity Level ($L_2$):} Comprises \textbf{Tool} and \textbf{Data} objects. Tools refer to functional units (e.g., Transformer blocks, external libraries, or specific algorithms), while Data represents I/O streams and intermediate features (e.g., feature maps, vectors, or raw text).
    \item \textbf{Component Level ($L_3$):} Constitutes the atomic visualization primitives within each object. We define three types of components: (1) \textit{Icon}: Functional identifiers (e.g., logos); (2) \textit{Text}: Descriptive strings (e.g., algorithm names, parameter descriptions); and (3) \textit{Image}: Pixel-level visuals (e.g., statistical charts or sample inputs).
\end{itemize}

\noindent \textbf{2. Data Schema and Constraints.} We serialize the graph structure using the JSON format. For any node $N$, the attributes are defined as follows:

\begin{itemize}
    \item \textbf{Metadata Attributes:} 
    The \texttt{type} field specifies the node category, where $\texttt{type} \in \{\texttt{module}, \texttt{tool}, \texttt{component-icon/text/image}\}$.
    Each node is assigned a globally unique identifier (\texttt{id}) and a semantic label (\texttt{name}). Note that while \texttt{name} can be reused across nodes, \texttt{id} must be unique.
    
    \item \textbf{Recursive Hierarchy:} 
    The \texttt{children} field defines the nesting structure. For $L_1$ and $L_2$ nodes, this field contains a list of child objects. For $L_3$ leaf nodes (components), this field stores the specific content payload (e.g., a text string, icon description, or image path).
    
    \item \textbf{Topological Connections:} 
    The \texttt{edges} field describes the logical data flow. To ensure modular decoupling, directed edges are strictly constrained to connect \textbf{sibling nodes} under the same parent node. 
    Each edge is defined as a tuple $\{id, name, sources, targets\}$, representing a directional connection from the source node(s) to the target node(s).
\end{itemize}
\begin{lstlisting}[language=json, caption={Example of the Graph JSON Data Structure.}, label={lst:json_example}]
{
    "type": "module",
    "id": "n1",
    "name": "Data Preprocessing Module",
    "children": [
        {
            "type": "tool",
            "id": "n2",
            "name": "Data Cleaning Tool",
            "children": [
                {
                    "type": "component-text",
                    "id": "n3",
                    "name": "Algorithm Name",
                    "children": "Outlier Detection&Handling"
                },
                {
                    "type": "component-icon",
                    "id": "n4",
                    "name": "Cleaning Icon",
                    "children": "broom_trash_icon.png"
                }
            ],
            "edges": [
                {
                  "sources": ["n2"], 
                  "targets": ["n5"], 
                  "id": "e1", 
                  "name": "Cleaned Data"
                }
            ]
        },
        {
            "type": "tool",
            "id": "n5",
            "name": "Data Standardization Tool",
            "children": [
                {
                    "type": "component-text",
                    "id": "n6",
                    "name": "Method",
                    "children": "Z-score Standardization"
                },
                {
                    "type": "component-image",
                    "id": "n7",
                    "name": "Example Chart",
                    "children": "path_to/image.png"
                }
            ]
        }
    ],
    "edges": [
        {
          "sources": ["n1"], 
          "targets": ["n5"], 
          "id": "e2", 
          "name": "Standardized Data"
        }
    ]
}
\end{lstlisting}

\vspace{2mm}

\subsubsection{Phase I: Semantic Parsing and Macro-Planning}
This phase aims to translate unstructured academic texts into a structured top-level topology. It involves two specific steps:

\begin{itemize}
    \item \textbf{Step 1: Information Extraction.} 
    The system deploys an \textit{Analyst Agent} to comprehend the input paper. The agent is instructed to extract core information across five dimensions: (1) overall task goals; (2) major modules and their responsibilities; (3) critical data flows (inputs/outputs/intermediate representations); (4) key algorithms or models; and (5) system constraints. This information is compressed into a structured system summary, serving as the contextual foundation for subsequent generation.

    \item \textbf{Step 2: Coarse-grained Topology Construction.} 
    Based on the extracted summary, an \textit{Architect Agent} constructs the skeletal graph. This step instantiates only the Root Node and the first-level ($L_1$) Module Nodes, establishing the macroscopic boundaries and directed connections between major modules without delving into internal details.
\end{itemize}

\subsubsection{Phase II: Distributed Generation and Global Alignment}
To handle the generation of complex system details, we adopt a ``Divide-and-Conquer'' strategy that transitions from parallel drafting to sequential refinement:

\begin{itemize}
    \item \textbf{Step 3: Parallel Sub-graph Instantiation.} 
    To ensure context isolation and generation efficiency, the system assigns an independent \textit{Designer Agent} to each $L_1$ module. These agents work in parallel to populate the internal $L_2$ Entity Level (Tools/Data) and $L_3$ Component Level. This phase generates local sub-graphs ($G_{sub}$) that guarantee local logical validity.

    \item \textbf{Step 4: Context-Aware Sequential Refinement.} 
    To address global consistency issues arising from parallel generation (e.g., ID conflicts or mismatched interfaces across modules), the system enters a serial revision mode. Local drafts are aggregated into a \textit{Global Context}. Each Designer Agent is sequentially activated to review and refine its responsible sub-graph under the global view, ensuring unique IDs and semantic alignment of I/O interfaces between modules.
\end{itemize}

\subsubsection{Phase III: Topological Regularization}
This phase enforces the constraints defined in the Hierarchical Graph Protocol using a \textbf{Neuro-Symbolic} approach:

\begin{itemize}
    \item \textbf{Step 5: Hybrid Constraint Enforcement.} 
    According to the protocol, directed edges are strictly prohibited from crossing module boundaries to connect internal components directly (preserving encapsulation). We employ a dual-check mechanism:
    \begin{enumerate}
        \item \textbf{Semantic Filtering (Neuro):} An LLM first reasons over the graph to identify and reroute obvious logical errors in cross-module connections.
        \item \textbf{Symbolic Pruning (Symbolic):} A deterministic program traverses all edges, strictly deleting any residual connections that violate topological constraints, ensuring the final graph structure is structurally legal.
    \end{enumerate}
\end{itemize}

\subsubsection{Phase IV: Adaptive Layout and Multi-modal Rendering}
The final phase transforms the logical graph structure into visually aesthetic and editable documents:

\begin{itemize}
    \item \textbf{Step 6: Content-Aware Layout Optimization.} 
    The system first dynamically calculates the initial size weights of nodes based on the length of their internal text content. Subsequently, it invokes the RectPacking algorithm from the \textbf{Eclipse Layout Kernel (ELK)}. This algorithm optimizes the geometric coordinates $(x, y, w, h)$ for each node, minimizing the canvas area while achieving a compact, non-overlapping layout.

    \item \textbf{Step 7: Generative Rendering and Vector Compilation.} 
    The system utilizes a Text-to-Image model to synthesize matching icons for each component. Finally, the \texttt{python-pptx} engine compiles all geometric information, text content, and image assets into editable PowerPoint slides and high-resolution images, completing the end-to-end generation.
\end{itemize}

\subsubsection{Prompt Engineering Details}
\label{sec:prompt-agent}

To enable the LLM to generate the specific hierarchical graph structure defined in our protocol, we designed a comprehensive system prompt. The prompt explicitly defines the role, the three-layer structure, and the strict JSON schema constraints. The full prompt is provided below in Figure \ref{fig:prompts_PaperInformationExtraction} and \ref{fig:prompt_agent_fig}

\begin{figure}[h]
    \begin{SplitBox}[%
        colback=violet!5,
        colframe=violet!75!black,
        colbacktitle=violet!80!black,
        fonttitle=\bfseries\sffamily,
        title=Prompt for Paper Information Extraction
        ]
        \textbf{< System >}\\ 
        \scriptsize \ttfamily % 使用小号打字机字体，容纳更多内容
            \textbf{\#\#\# Role:}\\
            You are an Expert in Understanding Academic System Architectures.
            
            \vspace{1mm}
            \textbf{\#\#\# Task Description}\\
            Given the original paper text (or excerpts), please \textbf{summarize} the following aspects point-by-point:
            
            \begin{itemize}[leftmargin=*, nosep]
                \item \textbf{System Name}
                \item \textbf{Task \& Overall Goal}
                \item \textbf{Main Modules/Stages \& Responsibilities}
                \item \textbf{Key Data/Information Flow} (Input $\rightarrow$ Process $\rightarrow$ Output)
                \item \textbf{Core Models/Algorithms \& Their Roles}
                \item \textbf{Key Constraints or Assumptions} (if applicable)
            \end{itemize}
            
            \vspace{1mm}
            \textbf{\#\#\# Requirements}\\
            \begin{itemize}[leftmargin=*, nosep]
                \item Language should be concise and well-organized.
                \item The structure must facilitate the subsequent conversion into a system architecture diagram.
            \end{itemize}
    
        \tcblower
        \textbf{< User >}\\
        \scriptsize \ttfamily
            Please analyze the following paper text.
        
            Paper Text: \codevar{\{paper\_content\}}
    \end{SplitBox}
\caption{The detailed Prompts used for Paper Information Extraction}
\label{fig:prompts_PaperInformationExtraction}
\end{figure}

\begin{figure*}[t]
    \centering
    % 调整框的样式
    \begin{SplitBox}[%
        colback=violet!5,
        colframe=violet!75!black,
        colbacktitle=violet!80!black,
        fonttitle=\bfseries\sffamily,
        title=Prompt for Graph Generation,
    ]
    \textbf{<System>}
        \scriptsize \ttfamily % 字体稍微再小一点，放进双栏更合适
        
        % 开始框内双栏
        \begin{multicols}{2} 
            
            \textbf{\#\#\# Role:}\\
            You are an expert Diagram Designer responsible for hierarchically designing and planning a system data flow diagram based on the given description text.
            
            \vspace{2mm}
            \textbf{\#\#\# Design Rules}\\
            The diagram follows a three-level hierarchical structure:
            \begin{itemize}[leftmargin=*, nosep]
                \item \textbf{Top Level (Module Layer):} Represents the overall system modules or phases (e.g., Intent Recognition, Data Preprocessing, Model Inference).
                \item \textbf{Middle Level (Tool/Data Layer):} 
                \begin{itemize}[nosep]
                    \item \textbf{Tool Objects:} Tools utilized within each module (e.g., GRU, Transformer, LLM interfaces).
                    \item \textbf{Data Objects:} Inputs, outputs, and intermediate results (e.g., feature maps, vectors).
                \end{itemize}
                \item \textbf{Bottom Level (Component Layer):} Concrete content contained within each module/tool/data object.
            \end{itemize}
            Directed edges can exist between two nodes belonging to the same parent node, determined by the data flow direction.
            
            \vspace{1mm}
            The Component Layer includes 3 types of components:
            \begin{enumerate}[leftmargin=*, nosep]
                \item \textbf{Icon Component (icon):} Icons assisting in explaining functions (e.g., LOGO).
                \item \textbf{Text Component (text):} Explanatory text (e.g., tool names, descriptions, I/O examples).
                \item \textbf{Image Component (image):} User-input images (e.g., charts).
            \end{enumerate}
            
            \vspace{2mm}
            \textbf{\#\#\# Design Requirements}\\
            \begin{itemize}[leftmargin=*, nosep]
                \item Focus primarily on the \textbf{logical layout}; do not consider spatial layout.
                \item Both Module and Tool Layers can directly contain components.
                \item For Component Layer nodes, include specific content: description for icons, body text for text, and file path for images.
                \item Ideally, every module should have an icon for illustration.
            \end{itemize}
            
            \vspace{2mm}
            \textbf{\#\#\# Output Format}\\
            Output in JSON format.
            
            \vspace{2mm}
            \textbf{\#\#\# Field Regulations}\\
            For an object:
            \begin{itemize}[leftmargin=*, nosep]
                \item Use \texttt{type} to distinguish hierarchy: \texttt{"module"}, \texttt{"tool"}, \texttt{"component-text/icon/image"}.
                \item Use \texttt{id} for a globally unique identifier string (e.g., "n1").
                \item Use \texttt{name} for the object's name.
                \item Use \texttt{children} to list child nodes. For Component Layer, it is a \textbf{single string} (content). For others, it is a list of objects.
                \item Use \texttt{edges} for connections \textbf{<between child objects>}.
                \begin{itemize}[nosep]
                    \item Format: \texttt{\{"sources": ["id"], "targets": ["id"], "id": "e\_id", "name": "label"\}}.
                    \item \texttt{sources} and \texttt{targets} are single-element lists.
                    \item Edges can only exist between objects under the same direct parent.
                \end{itemize}
            \end{itemize}
            
            \vspace{2mm}
            \textbf{Example:}\\
            (See Listing \ref{lst:json_example} for the JSON schema example.)
            
        \end{multicols} % 结束框内双栏
        \tcbline
        \normalsize \normalfont
        \textbf{<User> in Top-Level Design(Step2)}\\
        \vspace{2mm}
        \scriptsize \ttfamily
            Please design only the Top-Level Module Layer: Return a single JSON object representing the root object (type="module"). Its children must contain only Module Layer nodes (type="module"), along with the edges connecting these modules. Do not design Tool or Component Layers; strictly adhere to the aforementioned Output Format and Field Regulations.\\
            \codevar{\{User Input\}}\\
        \tcbline
        \normalsize \normalfont
        \textbf{<User> in Sub-graph Design(Step 3 and 4)}\\
        \vspace{2mm}
        \scriptsize \ttfamily
            Please perform sub-level design for the specified module. Generate only the Tool Layer (type="tool") and Component Layer (component-*) within this module. You may include edges to represent connections between child objects. Maintain the module ID; do not create cross-module edges. Strictly adhere to the output format.\\
            \begin{itemize}[leftmargin=*, nosep]
                \item Target Module:\codevar{\{module\_id\}}\\
                \item Original Requirement: \codevar{\{User Input\}}\\
                \item Context: 
                \begin{itemize}
                    \item "top\_design": \codevar{\{top design result in Step 2\}},\\
                    \item "other\_modules": \codevar{\{sibling designs of Step 3 (is empty during Step 3)\}},\\
                    \item "revision\_requirements": \codevar{\{extra revision requirements given during Step 4\}},\\
                \end{itemize}
            \end{itemize}
            Output Constraint: Return a single JSON object where type="module" and id matches the target module, containing its populated children and (optional) edges.\\
        \normalsize \normalfont
    \end{SplitBox}
    \caption{The detailed Prompts used to guide the LLM in generating structurally valid hierarchical graphs.}
    \label{fig:prompt_agent_fig}
\end{figure*}

% ----------------------------------------
\subsection{P2SA Benchmark Construction}
\label{ssec:benchmark_details}

To construct the Paper2SysArch Benchmark described the main paper, we implemented a standardized data collection and filtering pipeline. This section details the specific tools and parameters used.

\subsubsection{Data Parsing and Extraction}
We sourced papers from top-tier AI conferences (e.g., CVPR, ICCV, NeurIPS, ICLR) over the past five years. To handle the unstructured PDF data, we employed the \textbf{PyMuPDF} library:
\begin{itemize}
    \item \textbf{Image Extraction:} The pipeline iterates through PDF pages to extract all embedded bitmap resources, saving them as PNG files.
    \item \textbf{Text Association:} We extracted the full text and established page-level mapping between images and text. Specifically, the \textit{Abstract} and \textit{Figure Captions} were isolated to serve as contextual grounding for the subsequent filtering step.
\end{itemize}

\subsubsection{VLM-based Filtering and Cleaning}
To filter out non-architectural images (e.g., performance plots or qualitative results) and retain only genuine System Architecture Diagrams, we deployed a VLM-based discriminator:
\begin{itemize}
    \item \textbf{Discrimination Mechanism:} The extracted PNG images and the paper's abstract are fed into a VLM. The model is prompted to determine whether the image depicts the core system architecture or methodological pipeline of the paper.
    \item \textbf{Thresholding:} The model outputs a confidence score. We set a strict confidence threshold of \textbf{0.75}. Only images scoring above this value are candidates for retention.
    \item \textbf{Deduplication:} To ensure a one-to-one mapping for supervised training, we retained only the single highest-scoring architecture diagram for each paper.
\end{itemize}
Ultimately, combined with the manually curated test set, we established a dataset comprising 3,000 samples.

\subsection{P2SA Auto Evaluation System}
\label{ssec:eval_details}

This section provides the algorithmic implementation details and hyperparameter settings for the three-tier evaluation framework proposed in Sec. 3.4 of the main paper.

\subsubsection{Specialized Evaluation Agents}
\label{sec:eval_agents}

To address the issues of attentional dispersion and inconsistent instruction following inherent in general-purpose Vision-Language Models (VLMs) when processing information-dense diagrams, we propose a \textbf{Role-Based Multi-Agent Evaluation Framework}.

Instead of a coarse ``one-prompt-fits-all'' approach, we adopt the \textbf{``VLM-as-an-Agent''} paradigm. We decompose the complex task of system architecture evaluation into a series of atomic sub-tasks. Each agent represents a functional \textbf{encapsulation} of the VLM—constraining its broad multimodal capabilities into specialized expert discriminators via specific system prompts.

Specifically, we define five core agents to conduct orthogonal evaluations across structural, semantic, visual, and layout dimensions:

\paragraph{GraphExtractAgent (Topological Parsing Specialist):} 
This agent is responsible for abstracting the diagram's structure. It is instructed to ignore specific visual styles and focus solely on identifying nodes (entities) and edges (relations). It converts unstructured image pixels into a structured GraphJSON representation, providing a machine-readable intermediate format for subsequent semantic consistency calculations.

\paragraph{IconExamineAgent (Visual-Semantic Alignment Specialist):} 
This agent focuses on the alignment between visual elements and textual labels. It receives module definitions from the GraphJSON and locates the corresponding icon regions. Its core task is to judge whether the \textit{visual metaphor} of an icon accurately conveys the functional semantics of the module (e.g., checking if a ``Database'' module uses a storage bucket icon), thereby quantifying visual accuracy.

\paragraph{LayoutExamineAgent (Geometric Layout Auditor):} 
Acting as a visual QA specialist, this agent is dedicated to detecting geometric defects. It is specifically prompted to scan the image for ``violation patterns,'' such as unnecessary line crossings, text overflowing bounding boxes, and overlapping elements, to calculate layout rationality penalties.

\paragraph{SystemUnderstandAgent (Cognitive Simulator):} 
This agent simulates the high-level cognitive process of a \textit{human reader}. By combining the paper abstract (context) with the input image, it generates a natural language description of the system workflow. By comparing the agent's understanding of the ``Generated Diagram'' versus the ``Ground Truth,'' we can assess the \textit{information fidelity} at a macroscopic level.

\paragraph{TextLegibilityAgent (Low-level Perception Specialist):} 
Given that generative models often produce text artifacts, this agent is tasked with low-level visual perception. It focuses on detecting character distortion, blurring, or unreadable resolution, ensuring the usability of the diagram at the detail level.

Through this \textbf{divide-and-conquer} strategy, we focus the VLM's attention on single dimensions, significantly enhancing the robustness and interpretability of the automated evaluation.

\subsubsection{Semantic Layer Evaluation Details}
The assessment of semantic consistency relies on the accurate alignment between nodes in the Generated Graph ($N_g$) and the Ground Truth Graph ($N_{gt}$).

\noindent \textbf{1. Node Matching Algorithm.}
As defined in Eq. (1) of the main paper, the matching score aggregates multiple similarity features. In our implementation:
\begin{itemize}
    \item \textbf{Text Feature:} We utilize the \textbf{CLIP model} to compute the semantic cosine similarity between node labels.
    \item \textbf{Two-Stage Strategy:}
    \begin{itemize}
        \item \textit{Stage 1 (Anchor Matching):} This stage prioritizes CLIP-based text similarity and degree similarity to identify high-confidence anchor pairs.
        \item \textit{Stage 2 (Structure Propagation):} Building upon the anchors, we increase the weights of structural features (i.e., Neighbor Similarity and Ancestor Chain Similarity). This allows the system to recall nodes that perform identical structural roles but use paraphrased textual descriptions.
    \end{itemize}
\end{itemize}

\noindent \textbf{2. Scoring Mechanism.}
Instead of simple binary classification (Precision/Recall), we calculate a continuous \textbf{similarity score}:
\begin{itemize}
    \item \textbf{Node Consistency:} The average semantic similarity score of all matched node pairs.
    \item \textbf{Edge and Hierarchy Consistency:} Based on the aligned nodes, we calculate the similarity scores for edges (connectivity) and hierarchical relationships (parent-child containment) preserved in the generated graph.
\end{itemize}

\subsubsection{Layout Layer Evaluation Details}
Unlike rule-based SVG parsing, our layout evaluation is entirely \textbf{vision-based}, leveraging the perceptual capabilities of VLMs.
\begin{itemize}
    \item \textbf{Mechanism:} The \textit{LayoutExamineAgent} (a VLM) directly analyzes the generated PNG image to identify visual defects.
    \item \textbf{Penalty System:} The score starts at a perfect 1.0. For each detected instance of the following defects, \textbf{0.1 points} are deducted:
    \begin{itemize}
        \item \textit{Line Crossings:} Connection lines intersecting in non-node areas.
        \item \textit{Element Overlaps:} Unintended occlusion between nodes, icons, or text boxes.
        \item \textit{Text Overflows:} Text content extending beyond its container boundaries.
    \end{itemize}
\end{itemize}

\subsubsection{Visual Layer Evaluation Details}
This layer assesses aesthetics and information fidelity, also driven by VLM agents:

\begin{itemize}
    \item \textbf{Icon Relevance:} We do not employ separate segmentation models. Instead, the \textit{IconExamineAgent} (VLM) directly analyzes local image regions containing the icon and module name. It evaluates whether the visual metaphor of the icon accurately reflects the functional semantics of the module (e.g., checking if a "Data Cleaning" module uses a broom or filter icon).
    \item \textbf{System Understandability:} This measures information loss via semantic reconstruction. The VLM generates textual captions for both the generated diagram and the ground truth. We then calculate the semantic similarity between these two captions.
    \item \textbf{Text Legibility:} The \textit{TextLegibilityAgent} scans the image for artifacts specific to generative models, such as blurred, distorted, or unreadable characters.
\end{itemize}

\subsubsection{Prompt Engineering for Evaluation Agents}
\label{ssec:eval_prompts}

To instantiate the specialized evaluation capabilities of the VLM, we designed distinct system prompts for each of the five agents. These prompts explicitly define the agent's \textbf{persona} (e.g., ``Geometric Layout Auditor'' or ``Cognitive Simulator'') and outline the rigorous \textbf{adjudication criteria} (e.g., specific penalties for line crossings or blurriness). By conditioning the VLM with these structured instructions, we transform a general-purpose model into a suite of focused expert judges. The full system prompts for all evaluation agents are provided in Figures \ref{fig:prompts_GraphExtractAgent}-\ref{fig:prompts_TextLegibilityAgent}

\begin{figure}[h]
    \begin{SplitBox}[%
        colback=teal!5,
        colframe=teal!75!black,
        colbacktitle=teal!80!black,
        fonttitle=\bfseries\sffamily,
        title=Prompt for GraphExtractAgent
        ]
        \textbf{< System >}\\ 
        \scriptsize \ttfamily % 使用小号打字机字体，容纳更多内容
        
        \textbf{\#\#\# Role:}\\
        You are a Graph Processing Engineer responsible for extracting the topological structure from an input system architecture diagram and its description.
        
        \vspace{1mm}
        \textbf{\#\#\# Task Description}\\
        First, comprehend the image description. Then, based on your understanding of the diagram, deconstruct the input system architecture to extract its essence. Abstract it into a graph structure composed of nodes and edges, and represent it in JSON.
        
        \vspace{1mm}
        \textbf{\#\#\# Requirements}\\
        \begin{itemize}[leftmargin=*, nosep]
            \item Any semantic entity with independent content can be a node (e.g., a function, processing step, or I/O result). Do \textbf{not} over-split; for instance, a single sentence usually does not constitute a node. Content on lines is generally considered edge labels.
            \item \textbf{Granularity Control:} Node granularity should not be too fine; maintain consistency with the granularity of the \textbf{Image Description}!
            \item \textbf{Containment vs. Connection:}
            \begin{itemize}[nosep]
                \item If one node spatially encompasses another, it is a \textbf{containment} (parent-child) relationship.
                \item If nodes are linked by arrows, it is a \textbf{connection} relationship.
            \end{itemize}
            \item Judgment must rely on \textbf{semantics}, not just spatial layout.
        \end{itemize}
        
        \vspace{1mm}
        \textbf{\#\#\# Representation Rules}\\
        Top-level fields: \texttt{graph} and \texttt{explain}.
        \begin{enumerate}[leftmargin=*, nosep]
            \item \texttt{graph}: Contains \texttt{nodes} and \texttt{edges}.
            \begin{itemize}[nosep]
                \item \texttt{nodes}: List of objects. Each has \texttt{id} (unique), \texttt{name} (summarized by you), and \texttt{children} (list of child IDs; empty if leaf).
                \item \texttt{edges}: List of objects. Each has \texttt{source} (id), \texttt{target} (id), and \texttt{name} (summarized by you).
            \end{itemize}
            \item \texttt{explain}: Detailed explanation text.
            \begin{itemize}[nosep]
                \item Explain your understanding of the system's principles and purpose.
                \item Explain the content/function of each node.
                \item Explain the content/data flow of each edge.
            \end{itemize}
        \end{enumerate}
        
        \vspace{1mm}
        \textbf{\#\#\# Output Format}\\
        Output in JSON format wrapped in \textasciigrave\textasciigrave\textasciigrave json \textasciigrave\textasciigrave\textasciigrave.
        
        \vspace{1mm}
        \textbf{Example:}\\
        \textasciigrave\textasciigrave\textasciigrave json\{
            "graph": \{
                "nodes": [
                    \{ "id": "n0", "name": "Input Data", "children": [] \},
                    ...
                ],
                "edges": [        
                    \{"id": "e1", "source": "n0", "target": "n2", "name": "Data Input"\},
                    ...
                ]
            \}, 
        "explain": "..."
        \}\textasciigrave\textasciigrave\textasciigrave
        
        \tcblower
        
        \textbf{< User >}\\
        \scriptsize \ttfamily
        Please extract the graph structure based on the image content and the paper text.
        
        Paper Text: \codevar{\{paper\_text\}}
    \end{SplitBox}
\caption{The detailed Prompts used in GraphExtractAgent}
\label{fig:prompts_GraphExtractAgent}
\end{figure}

\begin{figure}[h]
    \begin{SplitBox}[%
        colback=teal!5,
        colframe=teal!75!black,
        colbacktitle=teal!80!black,
        fonttitle=\bfseries\sffamily,
        title=Prompt for IconExamineAgent
        ]
        \textbf{< System >}\\ 
        \scriptsize \ttfamily % 使用小号打字机字体，容纳更多内容
        \textbf{\#\#\# Role:}\\
        You are an Icon Inspection Engineer with a deep understanding of icon design and visual representation in system architecture diagrams.
        
        \vspace{1mm}
        \textbf{\#\#\# Task Description}\\
        Referencing the original paper text and the provided graph structure (nodes and edges), examine the icon for each module in the input image.
        \begin{itemize}[leftmargin=*, nosep]
            \item First, determine if a module has an icon.
            \item If \textbf{no icon} exists, return an empty string \texttt{""}.
            \item If an \textbf{icon exists}, return a textual description of the icon.
        \end{itemize}
        
        \vspace{1mm}
        \textbf{\#\#\# Requirements}\\
        \begin{itemize}[leftmargin=*, nosep]
            \item Your description must focus on the \textbf{visual appearance} of the icon itself.
            \item Do \textbf{not} be biased by the module's textual content or function name. The module information is provided \textbf{solely} to help you locate the module within the image, not to dictate your description of the icon.
        \end{itemize}
        
        \vspace{1mm}
        \textbf{\#\#\# Output Format}\\
        Output in JSON format wrapped in \texttt{```json}, containing no other text.
        \begin{itemize}[leftmargin=*, nosep]
            \item \textbf{Keys:} The \texttt{id} of each node in the graph structure.
            \item \textbf{Values:} The textual description of the icon.
        \end{itemize}
        
        \vspace{1mm}
        \textbf{Example:}\\
        \texttt{\textasciigrave\textasciigrave\textasciigrave json\{
            "n1": "A sophisticated robot typing text",
            "n2": "A Python language logo",
            "n3": "A database cylinder symbol"
        \}} \textasciigrave\textasciigrave\textasciigrave
        \tcblower
        
        \textbf{< User >}\\
        \scriptsize \ttfamily

        Please inspect the icons within the image based on the visual content, the paper text, and the graph structure.

        \vspace{1mm}
        Paper Text: \codevar{\{desc\}}\\
        Graph Structure: \codevar{\{graph\}}
    \end{SplitBox}
\caption{The detailed Prompts used in IconExamineAgent}
\label{fig:prompts_IconExamineAgent}
\end{figure}

\begin{figure}[h]
    \begin{SplitBox}[%
        colback=teal!5,
        colframe=teal!75!black,
        colbacktitle=teal!80!black,
        fonttitle=\bfseries\sffamily,
        title=Prompt for LayoutExamineAgent
        ]
        \textbf{< System >}\\ 
        \scriptsize \ttfamily % 使用小号打字机字体，容纳更多内容
        
        \textbf{\#\#\# Role:}\\
        You are a Diagram Layout Inspection Engineer capable of \textbf{meticulously} detecting layout anomalies in system architecture diagrams.
        
        \vspace{1mm}
        \textbf{\#\#\# Task Description}\\
        Examine the input diagram for the following layout issues:
        \begin{enumerate}[leftmargin=*, nosep]
            \item \textbf{Line Crossings:} Check for unreasonable intersections between connection edges or module bounding boxes.
            \begin{itemize}[nosep] \item \textit{Note: Intentional crossings designed for visual effect are excluded.} \end{itemize}
            
            \item \textbf{Element Overlaps:} Check for image/element overlaps that negatively impact aesthetics.
            \begin{itemize}[nosep] \item \textit{Note: Necessary stacking designs (e.g., card stacks) are excluded.} \end{itemize}
            
            \item \textbf{Text Overflows:} Check if text exceeds its bounding box or overlaps with other images or edges.
            \end{enumerate}
            
            \vspace{1mm}
            \textbf{\#\#\# Output Format}\\
            Output in JSON format wrapped in \texttt{\textasciigrave\textasciigrave\textasciigrave json \textasciigrave\textasciigrave\textasciigrave}, containing no other text.
            
            \vspace{1mm}
            \textbf{Example:}\\
            \texttt{\textasciigrave\textasciigrave\textasciigrave json\{
                "layout\_issues": [
                    \{ "type": "line\_crossing", "count": 2, "details": ["Desc 1...", "Desc 2..."] \},
                    \{ "type": "image\_overlap", "count": 1, "details": ["Desc..."] \},
                    \{ "type": "text\_overflow", "count": 1, "details": ["Desc..."] \}
                ]
            \}}
            \textasciigrave\textasciigrave\textasciigrave
        
        \tcblower
        \textbf{< User >}\\
        \scriptsize \ttfamily

        Please inspect the image for layout anomalies based on the visual content.

    \end{SplitBox}
\caption{The detailed Prompts used in LayoutExamineAgent}
\label{fig:prompts_LayoutExamineAgent}
\end{figure}

\begin{figure}[h]
    \begin{SplitBox}[%
        colback=teal!5,
        colframe=teal!75!black,
        colbacktitle=teal!80!black,
        fonttitle=\bfseries\sffamily,
        title=Prompt for SystemUnderstandAgent
        ]
        \textbf{< System >}\\ 
        \scriptsize \ttfamily % 使用小号打字机字体，容纳更多内容

            \textbf{\#\#\# Role:}\\
            You are a Human Researcher possessing a Master's degree level of cognition. You are responsible for understanding the operating principles of a system based on its architecture diagram and textual description.
            
            \vspace{1mm}
            \textbf{\#\#\# Task Description}\\
            Comprehend the input system architecture diagram and its accompanying text to provide an explanation of the system's operating principles.
            
            \vspace{1mm}
            \textbf{\#\#\# Requirements}\\
            \begin{itemize}[leftmargin=*, nosep]
                \item You must simulate \textbf{human cognitive levels and thinking patterns} as closely as possible when interpreting the image.
            \end{itemize}
            
            \vspace{1mm}
            \textbf{\#\#\# Output Format}\\
            Output in JSON format wrapped in \texttt{\textasciigrave\textasciigrave\textasciigrave json}, containing no other text.
            \begin{itemize}[leftmargin=*, nosep]
                \item Contains a single field: \texttt{system\_understanding} (string), representing your understanding of the system's operating principles.
            \end{itemize}
            
            \vspace{1mm}
            \textbf{Example:}\\
            \texttt{\{ "system\_understanding": "..." \}}\textasciigrave\textasciigrave\textasciigrave
        
        \tcblower
        \textbf{< User >}\\
        \scriptsize \ttfamily

        Please comprehend the system's operating principles based on the visual content, the paper text, and the graph structure.
    
        Paper Text: \codevar{\{desc\}}\\
        Graph Structure: \codevar{\{graph\}}
        
    \end{SplitBox}
\caption{The detailed Prompts used in SystemUnderstandAgent}
\label{fig:prompts_SystemUnderstandAgent}
\end{figure}

\begin{figure}[h]
    \begin{SplitBox}[%
        colback=teal!5,
        colframe=teal!75!black,
        colbacktitle=teal!80!black,
        fonttitle=\bfseries\sffamily,
        title=Prompt for TextLegibilityAgent
        ]
        \textbf{< System >}\\ 
        \scriptsize \ttfamily % 使用小号打字机字体，容纳更多内容

            \textbf{\#\#\# Role:}\\
            You are a Diagram Text Legibility Engineer capable of \textbf{meticulously} auditing text within diagrams for clarity issues word-for-word.
        
            \vspace{1mm}
            \textbf{\#\#\# Task Description}\\
            Read the input system architecture diagram verbatim and inspect it for issues regarding blurriness, incompleteness, and semantic ambiguity.
        
            \vspace{1mm}
            \textbf{\#\#\# Requirements}\\
            \begin{itemize}[leftmargin=*, nosep]
                \item \textbf{Blurriness}: Includes text distortion, non-existent character structures (hallucinations), or text that is mashed into a blur.
                \item \textbf{Incompleteness}: Includes text ending abruptly in inappropriate places, or text artifacts appearing where they should not exist.
                \item \textbf{Semantic Ambiguity}: Includes unintelligible text or text that violates conventional semantics (gibberish).
            \end{itemize}
        
            \vspace{1mm}
            \textbf{\#\#\# Output Format}\\
            Output in JSON format wrapped in \texttt{\textasciigrave\textasciigrave\textasciigrave json}, containing no other text.
            \begin{itemize}[leftmargin=*, nosep]
                \item The top-level field is \texttt{text\_legibility\_issues}, which is a list of dictionaries. Each dictionary contains \texttt{type}, \texttt{count}, and \texttt{details}.
                \item \texttt{type} values: "Blurry", "Incomplete", "Ambiguous".
            \end{itemize}
        
            \vspace{1mm}
            \textbf{Example:}\\
            \texttt{\{
                "text\_legibility\_issues": [
                    \{ "type": "Blurry", "count": 2, "details": ["Desc 1...", "Desc 2..."] \},
                    \{ "type": "Incomplete", "count": 1, "details": ["Desc..."] \},
                    \{ "type": "Ambiguous", "count": 1, "details": ["Desc..."] \}
                ]
            \}}\textasciigrave\textasciigrave\textasciigrave
        
        \tcblower
        
        \textbf{< User >}\\
        \scriptsize \ttfamily
            Please inspect the text in the diagram for issues regarding blurriness, incompleteness, and semantic ambiguity based on the visual content.

    \end{SplitBox}
\caption{The detailed Prompts used in TextLegibilityAgent}
\label{fig:prompts_TextLegibilityAgent}
\end{figure}

\clearpage
\section{Detailed Case Studies }
\label{sec:case_study}

To provide a granular, end-to-end demonstration of our framework, we include a detailed case study in the supplementary archive (\texttt{supp.zip}). This case study traces a single research paper, ``Browsing Like Human,'' through both our generation and evaluation pipelines, offering a microscopic view of the system's internal mechanisms.

The study is organized into two directories: \texttt{Generation Case} and \texttt{Evaluation Case}.

\vspace{2mm}
\noindent \textbf{1. Generation Task (\texttt{/Generation Case})}

This directory contains all artifacts related to the generation process:
\begin{itemize}[leftmargin=*, nosep]
    \item \texttt{Browsing Like Human...pdf}: The original source paper used as input.
    \item \texttt{graph\_design\_detailed\_results.txt}: A comprehensive log detailing the multi-agent reasoning process, from the initial analysis by the \textit{Analyst Agent} to the final refined GraphJSON produced by the collaborative workflow. This file is crucial for understanding the step-by-step construction logic.
    \item \texttt{output figure.png}: The final rendered diagram as a static image.
    \item \texttt{output figure.pptx}: The same diagram in a fully editable PowerPoint format, demonstrating a key output of our system.
\end{itemize}

\noindent \textbf{2. Evaluation Task (\texttt{/Evaluation Case})}

This directory contains the corresponding materials for evaluating the generated output:
\begin{itemize}[leftmargin=*, nosep]
    \item \texttt{GroundTruth.png}: The ground-truth diagram extracted from the source paper, serving as the benchmark for comparison.
    \item \texttt{eval\_detailed\_results.txt}: The full output log from our automated evaluation framework. It shows the step-by-step assessment across all three dimensions (Semantic, Layout, and Visual) and provides detailed justifications for the final scores.
\end{itemize}

\section{Additional Dataset Statistics and Stability Analysis}
\label{sec:additional_stats}

In addition to the test set analysis presented in the main paper, this section provides a comprehensive statistical overview of the full training dataset and evaluates the robustness of our automated evaluation framework.

\subsection{Full Dataset Overview}
While Sec. 3.3 focused on the curated 108-sample test set, here we present the statistics for the complete \textbf{Paper2SysArch-3k} dataset to demonstrate its scale and diversity.

\begin{figure*}[t] % [t]表示尽量放在页顶，figure*表示跨双栏通栏显示
    \centering
    
    % --- 第一张图 (a) ---
    \begin{subfigure}{0.32\linewidth}
        \centering
        % width=\linewidth 表示填满这个子图的0.32宽度
        \includegraphics[width=\linewidth]{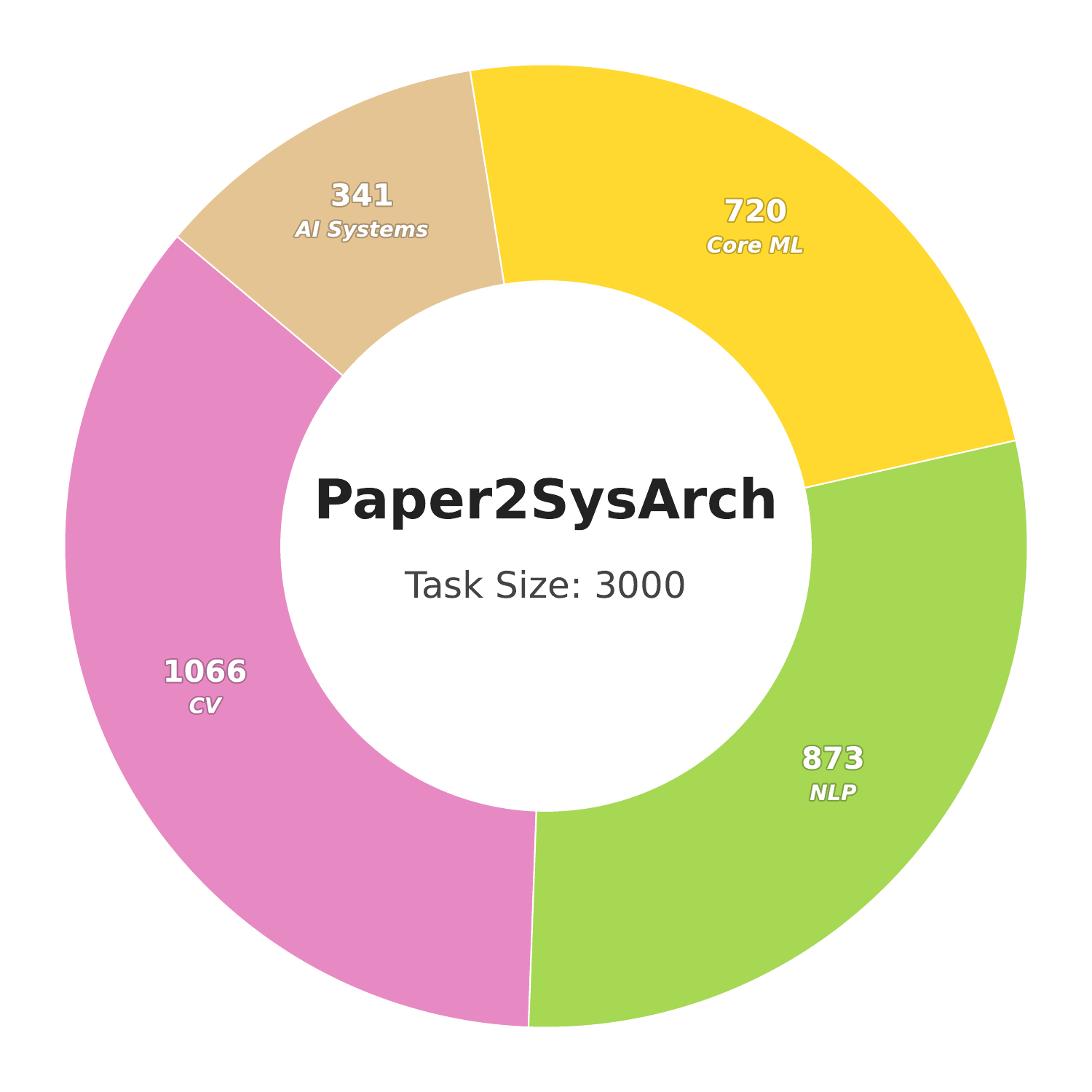} 
        \caption{Research Domains} % 这里会自动生成 (a)
        \label{fig:stats_domain}
    \end{subfigure}
    \hfill % 【关键】这个命令会把图片向两边推，产生均匀间隙
    % --- 第二张图 (b) ---
    \begin{subfigure}{0.32\linewidth}
        \centering
        \includegraphics[width=\linewidth]{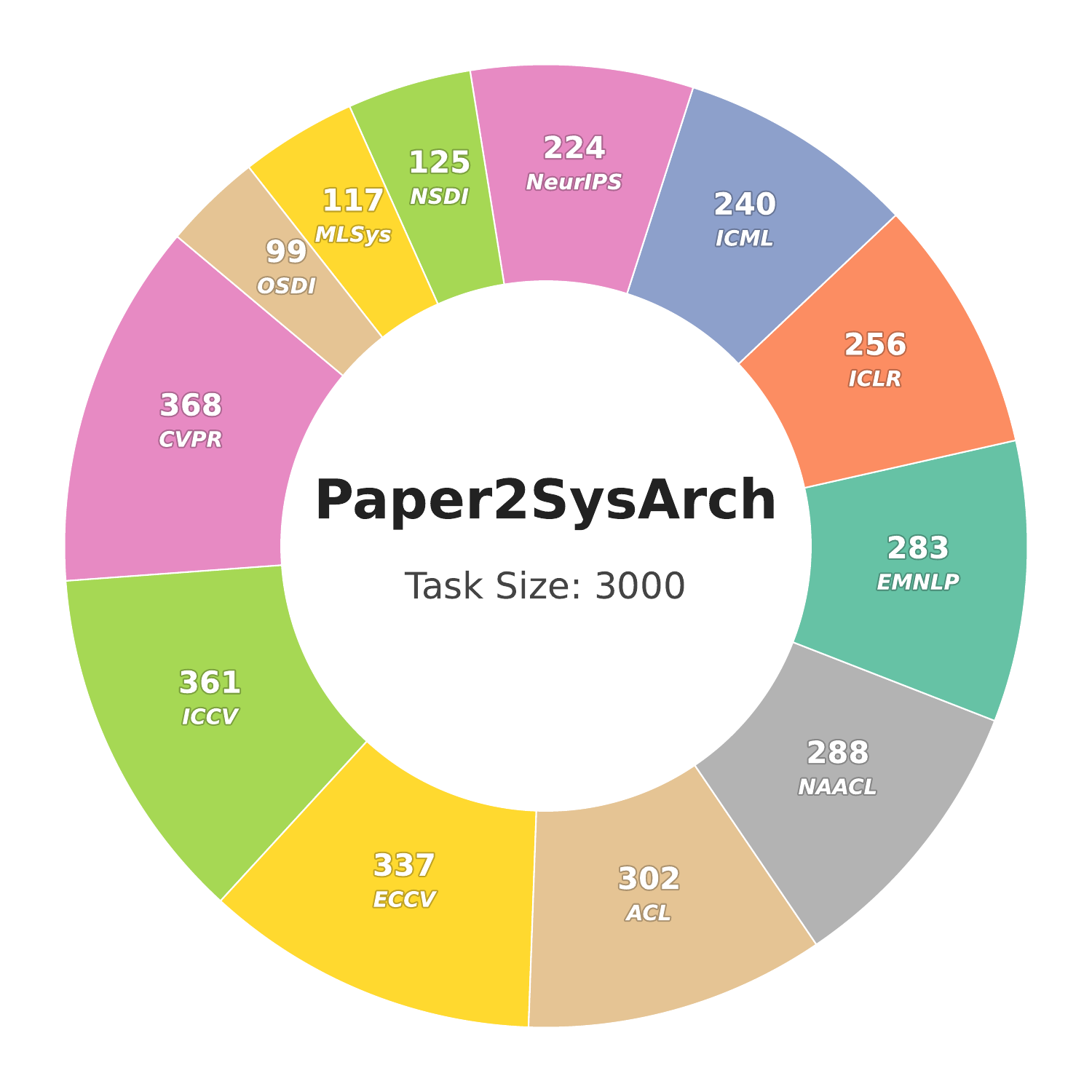}
        \caption{Source Conferences} % 这里会自动生成 (b)
        \label{fig:stats_conf}
    \end{subfigure}
    \hfill % 【关键】
    % --- 第三张图 (c) ---
    \begin{subfigure}{0.32\linewidth}
        \centering
        \includegraphics[width=\linewidth]{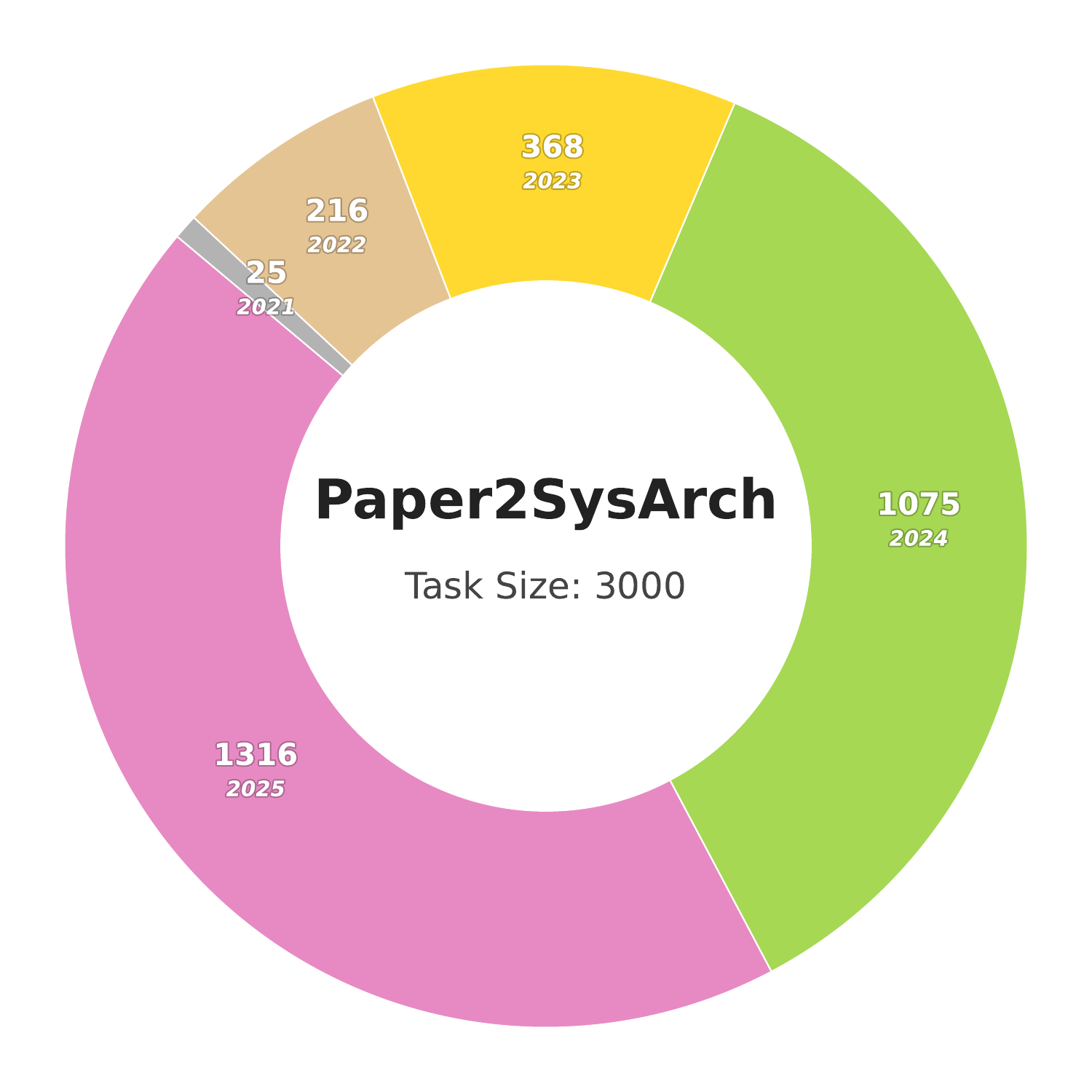}
        \caption{Publication Years} % 这里会自动生成 (c)
        \label{fig:stats_year}
    \end{subfigure}

    \caption{Statistical distribution of the full Paper2SysArch dataset (3,000 samples) across (a) Research Domains, (b) Source Conferences (12 venues), and (c) Publication Years.}
    \label{fig:dataset_stats}
\end{figure*}

\begin{itemize}
    \item \textbf{Domain Distribution:} As shown in Figure~\ref{fig:dataset_stats}(a), the dataset covers four core AI domains. Computer Vision (CV) accounts for the largest share (1066), followed by NLP (873) and Core ML (720), with a significant portion dedicated to AI Systems (341).
    
    \item \textbf{Conference Source:} Figure~\ref{fig:dataset_stats}(b) illustrates that the data originates from \textbf{12 top-tier conferences}. This includes premier venues in Vision (CVPR, ICCV, ECCV), Language (ACL, NAACL, EMNLP), Machine Learning (NeurIPS, ICML, ICLR), and Systems (OSDI, NSDI, MLSys), ensuring authoritative and diverse data sources.
    
    \item \textbf{Temporal Coverage:} As depicted in Figure~\ref{fig:dataset_stats}(c), priority is given to recent research, with papers from 2024 (1075) and 2025 (1316) constituting the majority. This ensures the benchmark reflects state-of-the-art architectural trends.
\end{itemize}

To facilitate a direct inspection of data quality, we provide a \textbf{Dataset Sample} in the supplementary archive (\texttt{/dataset\_sample}), containing representative paper-diagram pairs from each conference.

\subsection{Stability Analysis of Evaluation System}
Given the involvement of VLM-based agents in our evaluation framework, the potential for \textbf{stochastic fluctuations in scoring is a critical aspect that warrants investigation}. To verify the consistency and reproducibility of the metric, we conducted a repeatability experiment.

\textbf{Setup:} We randomly selected 10 samples (IDs 1-10) from the test set and executed the full evaluation pipeline 5 times for each sample independently, keeping all parameters constant.

\begin{figure*}[h]
    \centering
    \includegraphics[width=1.0\linewidth]{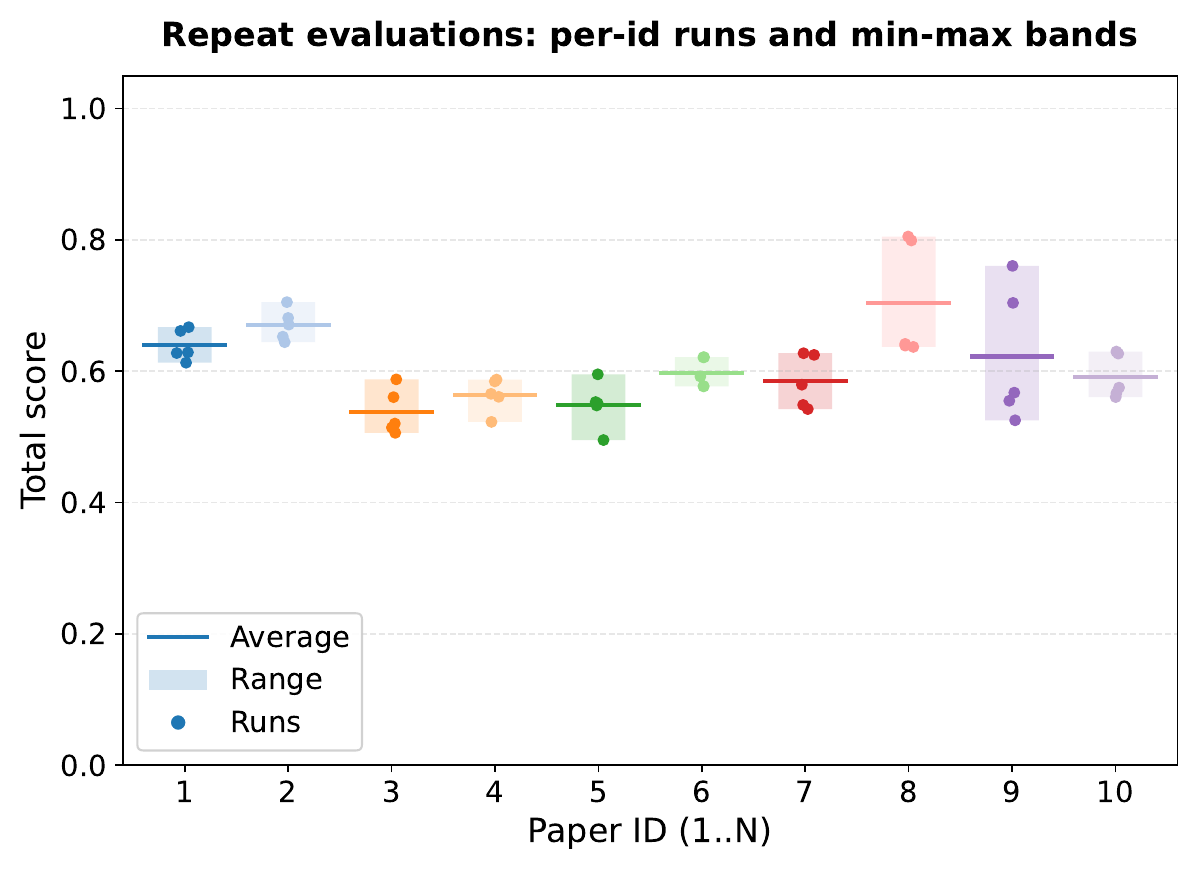} % 替换成你的文件名
    \caption{Stability analysis of the evaluation framework. We performed 5 repeat runs for 10 randomly selected paper IDs. The plot shows the individual run scores (dots), the average score (solid line), and the min-max range (shaded area), demonstrating the consistency of our metric.}
    \label{fig:stability}
\end{figure*}

\textbf{Results:} As shown in Figure~\ref{fig:stability}, we plot the individual scores (dots), average scores (lines), and min-max ranges (shaded bands) for each ID. The results demonstrate high stability, with most samples showing minimal fluctuation. Even in cases with slight variance (e.g., ID 8), the deviation remains within a narrow, acceptable range. This confirms that the Paper2SysArch evaluation framework produces consistent and reliable assessments.

\vspace{2mm}

\end{document}